\pdfoutput=1

\documentclass[11pt]{article}

\usepackage[preprint]{acl}

\usepackage{times}
\usepackage{makecell}
\usepackage{adjustbox}
\usepackage{latexsym}
\usepackage{times}
\usepackage{adjustbox}
\usepackage{float}
\usepackage{diagbox}
\usepackage{latexsym}
\usepackage{amsmath}
\usepackage{ulem}
\usepackage{amssymb}
\usepackage{graphicx}
\usepackage{booktabs}
\usepackage{multirow}
\usepackage{url}
\usepackage{tcolorbox}
\usepackage{mathabx}
\usepackage[table]{xcolor}
\definecolor{markcolor}{RGB}{220,224,228}
\definecolor{hiscolor}{RGB}{220,224,228}
\definecolor{applegreen}{rgb}{0.55, 0.71, 0.0}
\definecolor{downred}{RGB}{255, 136, 132}
\definecolor{rolecolor_1}{RGB}{255,247,172}
\definecolor{rolecolor_2}{RGB}{221,241,243}
\definecolor{rolecolor_3}{RGB}{236,244,221}
\definecolor{intercolor_1}{RGB}{255,230,230}
\definecolor{intercolor_2}{RGB}{221,242,247}
\definecolor{bestcolor}{RGB}{142,139,254}
\newcounter{mytcolorbox}

\usepackage[T1]{fontenc}

\usepackage[utf8]{inputenc}

\usepackage{microtype}

\usepackage{inconsolata}

\usepackage{graphicx}

\title{Concept than Document: Context Compression via \\ AMR-based Conceptual Entropy}

\author{Kaize Shi $^{\heartsuit}$, Xueyao Sun$^{\clubsuit, \spadesuit}$, Xiaohui Tao $^{\heartsuit}$, Lin Li $^{\bigstar}$, Qika Lin $^{\blacksquare}$, Guandong Xu$^{\diamondsuit}$\\ \\
  $^{\heartsuit}$ University of Southern Queensland, 
  $^{\clubsuit}$ University of Technology Sydney \\
  $^{\spadesuit}$ The Hong Kong Polytechnic University, 
  $^{\bigstar}$ Wuhan University of Technology \\
  $^{\blacksquare}$ National University of Singapore, 
  $^{\diamondsuit}$ The Education University of Hong Kong
}

\begin{document}
\maketitle
\begin{abstract}
Large Language Models (LLMs) face information overload when handling long contexts, particularly in Retrieval-Augmented Generation (RAG) where extensive supporting documents often introduce redundant content. This issue not only weakens reasoning accuracy but also increases computational overhead. We propose an unsupervised context compression framework that exploits Abstract Meaning Representation (AMR) graphs to preserve semantically essential information while filtering out irrelevant text. By quantifying node-level entropy within AMR graphs, our method estimates the conceptual importance of each node, enabling the retention of core semantics. Specifically, we construct AMR graphs from raw contexts, compute the conceptual entropy of each node, and screen significant informative nodes to form a condensed and semantically focused context than raw documents. Experiments on the PopQA and EntityQuestions datasets show that our method outperforms vanilla and other baselines, achieving higher accuracy while substantially reducing context length. To the best of our knowledge, this is the first work introducing AMR-based conceptual entropy for context compression, demonstrating the potential of stable linguistic features in context engineering.
\end{abstract}

\section{Introduction}

Large Language Models (LLMs) are increasingly equipped with mechanisms to incorporate long contexts, allowing them to leverage external information beyond their training data~\cite{lewis2020retrieval, karpukhin2020dense}. However, as the context length grows, LLMs often struggle to effectively identify and utilize truly relevant information, leading to performance degradation and inefficiency. This challenge, reflecting the trade-off between retrieval recall and precision, becomes particularly acute in scenarios such as Retrieval-Augmented Generation (RAG), where the inclusion of more retrieved documents raises the chance of accessing useful knowledge but simultaneously introduces overwhelming amounts of irrelevant text obscure the key facts~\cite{shi2023replug, jin2025hierarchical}.

\begin{figure}
\centering
\includegraphics[width=1\linewidth]{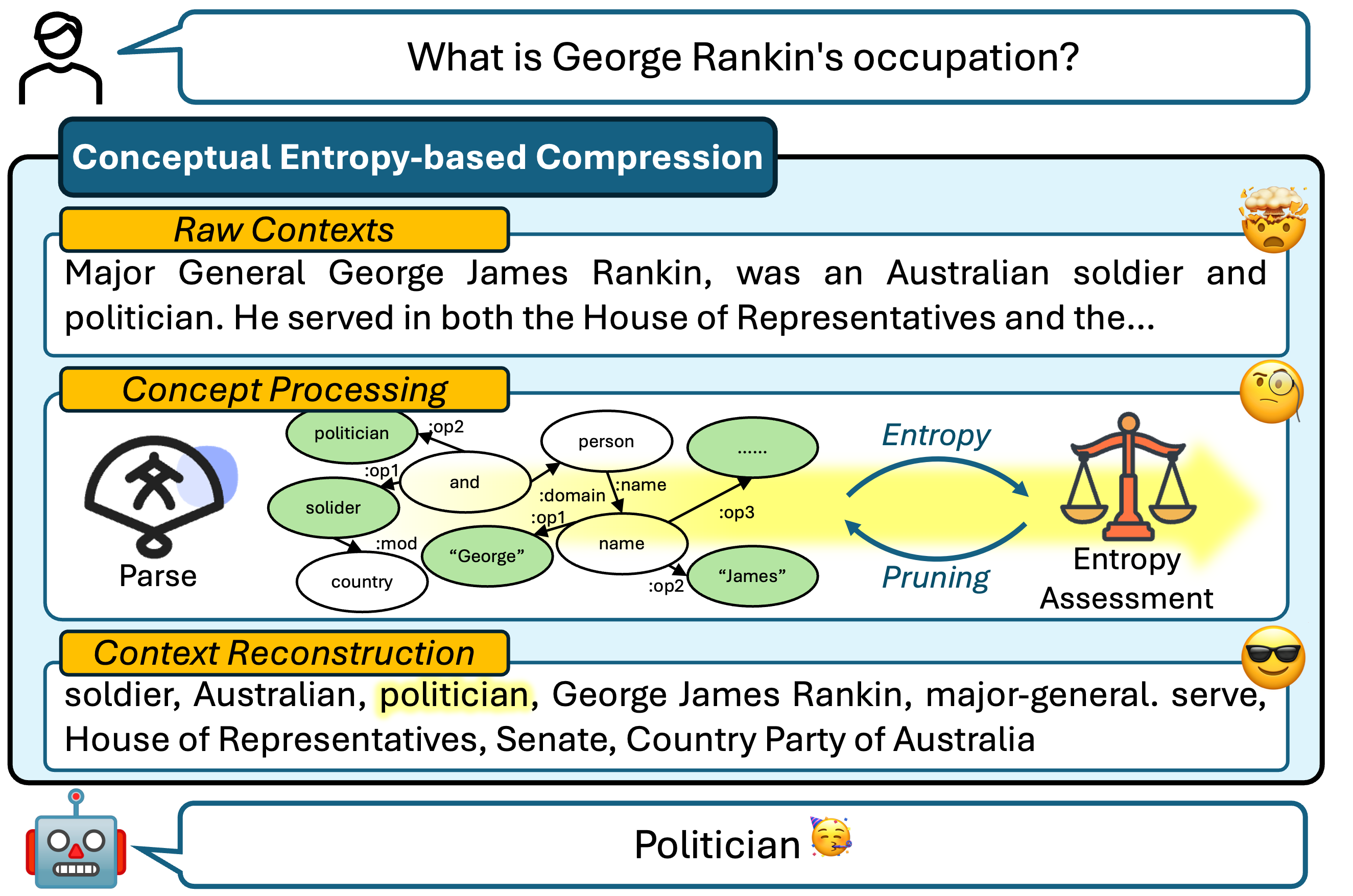}
\caption{Long retrieved documents contain much irrelevant content; our method keeps only key AMR-based concepts to form a semantically focused context.}
\label{fig:toy_example}
\end{figure}

Context engineering has therefore become an effective strategy for enhancing the quality and efficiency of long-context utilization, aiming to distill essential information while reducing noise and redundancy~\cite{mei2025survey}. Existing approaches primarily focus on lexical or surface-level features for information filtering~\cite{xu2023recomp, cheng2024xrag}. While these methods work well for certain queries, they may struggle with capturing complex semantic relationships and preserving factually important information. Moreover, traditional compression techniques may inadvertently remove crucial supporting evidence while retaining superficially relevant but semantically vacuous content.

To address the aforementioned limitations, we propose a novel context compression method that leverages Abstract Meaning Representation (AMR)~\cite{banarescu-etal-2013-abstract} to identify and preserve semantically essential information. AMR graphs provide a structured representation that abstracts away from surface syntactic variations while retaining core semantic content~\cite{chen2025semantic}. Concepts assuming diverse semantic roles across contexts naturally carry more informative value for inference~\cite{kuhn2023semantic}, which can be quantified as higher information entropy in the role distribution of concept nodes~\cite{nguyen-etal-2025-beyond}. Moreover, cognitive studies suggest that the human brain can automatically reconstruct scenarios implied by core concepts through pre-learned semantic knowledge~\cite{binder2009semantic, horikawa2025mind}, and LLMs exhibit a similar capacity for concept-based scene understanding, providing theoretical support for prioritizing semantically fundamental concepts during reasoning~\cite{du2025human}.

Building on this foundation, our method constructs AMR graphs from retrieved contexts, capturing both entities and their interrelations in a structured semantic form. For each concept node, we compute information entropy to assess its semantic contribution, considering its role and relational context. We then apply significance testing to identify truly informative nodes, which form the backbone for reconstructing a compressed context that preserves critical semantic information while discarding redundant or irrelevant content. To mitigate potential distortion caused by AMR’s abstraction from surface realization, the selected concepts are restored to their original textual expressions in the source contexts, ensuring factural consistency and maintaining semantic clarity in the reconstructed compressed context for reasoning.

We evaluate our method on two challenging knowledge-intensive Q\&A benchmarks, PopQA~\cite{mallen-etal-2023-trust} and EntityQuestions~\cite{sciavolino-etal-2021-simple}, which require reasoning over long-context factual evidence retrieved from external sources. Experimental results show substantial performance gains over vanilla RAG and existing context compression baselines, with particularly strong improvements on instances involving long supporting documents. These findings support our hypothesis that AMR-based entropy filtering effectively isolates core semantic content while removing redundant information. The main contributions of this work are summarized as follows:

\begin{itemize}
\item We propose a novel unsupervised context compression framework that leverages AMR to identify and preserve core semantic information while filtering redundant content.

\item Extensive experiments demonstrate that the proposed method outperforms vanilla and other compression baselines by maintaining robust semantic core preservation.

\item The method achieves reductions in context length and latency while preserving semantic integrity, offering a linguistically empowered framework for context engineering.
\end{itemize}

\section{Related Work}

\subsection{Context Engineering}

Context engineering has become a key strategy for managing and structuring information in LLM workflows~\cite{mei2025survey, verma2024contextual, shi2024compressing}. Early approaches selected relevant sentences or passages based on lexical similarity~\cite{hwang2024exit}, while later methods used neural models to reorganize retrieved contexts~\cite{xu2023recomp, liu2024towards}. Recent work examines learned context engineering techniques that optimize representations for downstream tasks. \citet{jiang2023longllmlingua} uses instruction tuning to refine contexts while preserving task-relevant information. Selective-Context~\cite{li2023compressing} applies attention mechanisms to highlight critical segments. \citet{jin2025sara} emphasizes semantic integrity in engineered contexts, integrating natural language spans and semantic vectors to support dynamic evidence selection and improve answer quality.

\subsection{AMR-enhanced Large Language Models}

Abstract Meaning Representation provides a structured formalism that abstracts away from syntactic variations, making it suitable for cross-lingual and cross-domain applications~\cite{wein-opitz-2024-survey}. Recent AMR parsing advances have made it practical to construct high-quality graphs from context~\cite{bevilacqua2021one, zhou-etal-2021-amr}, enabling applications across NLP tasks\cite{li-etal-2021-addressing-semantic, liu-etal-2015-toward, song-etal-2019-semantic}. With the rise of LLMs, researchers have explored using AMR for semantic enhancement. Recent studies examined AMR-driven chain-of-thought prompting, showing that structured semantic representations can improve LLM performance across tasks~\cite{jin-etal-2024-analyzing}. Other work has integrated AMR into LLM frameworks through structured representation methods, though challenges remain in aligning AMR's graph structure with sequential processing~\cite{zhang2025sr}. AMR nodes encode high-entropy semantic abstractions that capture rich conceptual information, enabling structured context engineering with more effective information use.

\subsection{Information Theory in LLMs}

Information-theoretic measures have become increasingly important in the era of LLMs, providing principled tools to understand and improve model behavior~\cite{wang2025learning}. LLMs have leveraged such analyses for interpretation and optimization~\cite{NEURIPS2024_10c456d2}. For instance, entropy-based selection of demonstration examples has been shown to enhance the performance of CoT prompting~\cite{zhou-etal-2023-inform}. Beyond prompting, information-theoretic approaches have been applied to model compression, knowledge distillation, and efficient fine-tuning~\cite{yin2024entropy, mao2024parse}. These studies illustrate an emerging trend in which information theory provides both theoretical insights and practical tools for working with LLMs~\cite{agarwal2025unreasonable}. In this work, we integrate graphical information-theoretic principles of AMR, leveraging high-entropy nodes as concise and informative representations of long contexts.

\section{Methodology}

\begin{figure*}
\centering
\includegraphics[width=1\linewidth]{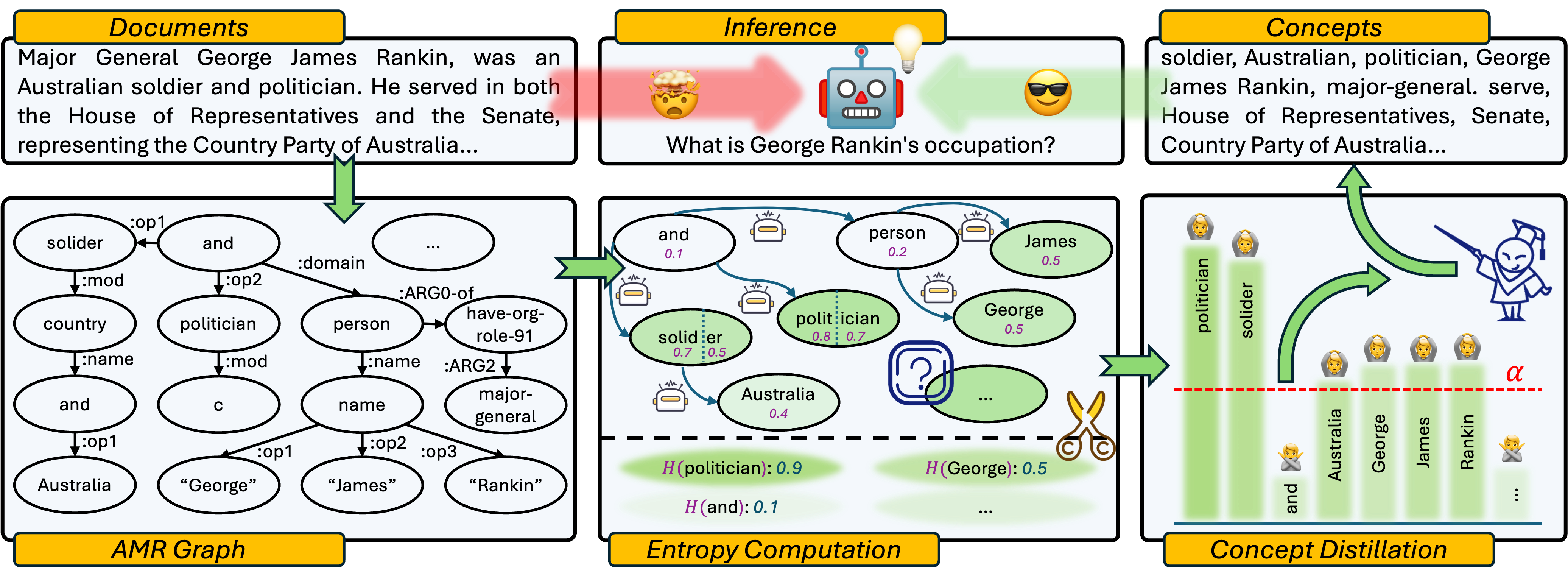}
\caption{The conceptual entropy-based workflow converts the sparse context in raw supporting documents into condensed AMR-based concepts, forming a compact semantic representation for LLMs inference.}
\label{fig:framework}
\end{figure*}

\subsection{Problem Formulation}

The framework for transferring the context in raw documents to condensed concepts is as Figure~\ref{fig:framework}. Given a query $Q$ and a set of retrieved documents $D = \{d_1, d_2, ..., d_n\}$ with corresponding correct answers $A = \{a_1, a_2, ..., a_m\}$, our objective is to generate a compressed context $C'$ that preserves the most semantically informative concepts essential for answering $Q$ to yield $a_j \in A$, while substantially reducing the overall context length.

To create a controlled experiment that focuses exclusively on the impact of core concepts within the context on answer accuracy, we retain only documents that contain correct answers. This controlled setting enables us to isolate how our compression method affects the preservation of essential contextual information by eliminating interference from irrelevant documents. The hypothesis can be formalized as: $\forall d_i \in D$, $\exists a_j \in A$ such that $a_j \in d_i$.

Formally, we aim to learn a compression function $f(D) \rightarrow C'$ such that:
\begin{equation}
\label{eq_formulation}
{Acc}(q, C') \gtrsim {Acc}(q, D) \text{ and } |C'| \ll |D|
\end{equation}
where $C' \subseteq D$, $|C'|$ and $|D|$ are the lengths of the compressed and original contexts, respectively.

\subsection{AMR Graph Construction}

For each document $d_i \in D$, we construct it to the sentence-level AMR graphs with an mBart-based parser~\footnote{\url{https://github.com/BramVanroy/multilingual-text-to-amr}} trained in the AMR 3.0 corpus~\footnote{\url{https://catalog.ldc.upenn.edu/LDC2020T02}} to address potential multilingual concerns. Let $G_i = (V_i, E_i)$ denote the AMR graph for document $d_i$, where $V_i$ represents the set of concept nodes and $E_i$ represents the semantic relations between concepts. Each concept node $v \in V_i$ corresponds to a semantic concept (e.g., entities, predicates, or modifiers) and is associated with its textual realization in the raw document. The edges in $E_i$ represent semantic relationships such as agent-of (ARG0), patient-of (ARG1), and various semantic roles.

Our approach is grounded in the cognitive hypothesis that both human comprehension and LLM inference can effectively reconstruct semantic scenarios from discrete informative concepts without explicit relational encoding~\cite{xu2025large, fedorenko2024language, rogers2004structure, wit1999linguistic}. This principle suggests that intelligent systems possess inherent capabilities to infer implicit relationships between concepts based on their learned background knowledge and contextual co-occurrence patterns~\cite{NEURIPS2020_1457c0d6, cao-etal-2023-unnatural, suresh-etal-2023-conceptual}. Building on these foundations, we keep the concept nodes $V_i$ and discard the explicit $E_i$ in each $G_i$. This design ensures that the compressed context consists of discrete semantic concepts, avoiding the introduction of artificial relational symbols that may interfere with the LLM's pre-trained language understanding capabilities while leveraging the model's intrinsic ability in concept-based scenario reconstruction.

\subsection{Information Entropy Computation}

To identify the most informative concepts within each AMR graph, we employ an information-theoretic approach based on token-level perplexity measurements. For each concept node $v \in V_i$, we calculate its information entropy by leveraging the AMR generation model's uncertainty when predicting the concept token sequence.

Given the AMR parsing model $M$ with parameters $\theta$, we obtain the probability distribution over the vocabulary for each token position in the AMR linearization. However, modern tokenizers decompose words into subword units, requiring the aggregation to obtain concept-level entropy scores. For a concept $v$ that corresponds to a complete word-level representation in $d_i$, the tokenizer may decompose it into the subword tokens $v = [s_1, s_2, ..., s_m], m \geq 0$. We compute the token-level entropy for each subword as:
\begin{equation}
E(s_j) = \exp\left(-\log P_\theta(s_j | s_{<j}, G_i)\right)
\end{equation}
where $s_{<j}$ denotes the preceding tokens within the same concept. We aggregate token-level entropies into a concept-level entropy score as Eq.~\ref{eq_Hv}. Specifically, we identify concept boundaries by tracking tokens that begin with the special prefix "Ġ" and accumulate entropy scores for all $s_j$ belonging to the same conceptual unit. This aggregation strategy ensures that concepts composed of multiple subword tokens are not artificially penalized relative to single-token concepts. This alignment provides a balanced representation of the model's uncertainty across all subword components of a concept.
\begin{equation}
\label{eq_Hv}
H(v) = \frac{1}{m}\sum_{j=1}^{m} E(s_j)
\end{equation}

Compared to token-level entropy in linear text, computing entropy over AMR concept nodes leverages semantic structure to more precisely estimate informational content. High-entropy nodes often represent content-specific, less redundant meanings, thus providing more discriminative signals for downstream reasoning. This enables the compression process to highlight semantically rich units that may be obscured in the surface text.

\subsection{Concept Distillation}

The supporting document set $D$ can be conceptualized as a coherent descriptive scenario corresponding to query $Q$, within which genuinely informative concepts can be identified through their statistically significant entropy deviations. Concepts exhibiting higher entropy relative to the general nodes carry more discriminative information and are thus more valuable for answering the query. For each $d_i \in D$ with concept entropy $\{H(v_1), H(v_2), ..., H(v_{|V_i|})\}$, we perform a one-sample t-test to identify concepts with significantly higher information than the population mean:
\begin{equation}
t_{stat}(v_j) = \frac{H(v_j) - \bar{H}}{\frac{s}{\sqrt{n}}}
\end{equation}
where $\bar{H}$ is the sample mean entropy, $s$ is the sample standard deviation, and $n = |V_i|$. We compute the corresponding p-value using the t-distribution with $n-1$ degrees of freedom:
\begin{equation}
p(v_j) = 2 \times (1 - F_t(|t_{stat}(v_j)|, n-1))
\end{equation}
where $F_t$ is the cumulative distribution function. We then screen out concepts whose p-values satisfy $p(v_j) < \alpha$ as statistically significant high-information concepts. Our goal is not to identify only the most informative concepts, but rather to eliminate overly generic ones while preserving a relative conceptual basis for LLMs to infer the document's semantics. Considering the empirical validation of LLMs' inference, we adopt a relaxed threshold, $\alpha = 0.3$. This setting prevents the over-pruning of moderately informative concepts, thereby ensuring that the retained set includes contextual signals. The ablation study to verify the different $\alpha$ settings is in Section~\ref{sec:ablation}.

\subsection{Context Compression and Reconstruction}

The final compressed context $C'$ is constructed by aggregating the concepts with significant entropy across all documents in $D$. For each document $d_i$, let $V_i = \{v \in V_i : p(v) < \alpha\}$ denote the set of statistically significant concepts. For each $c_i \in C'$, the compressed representation for document $d_i$ is:
\begin{equation}
c_i = \bigodot_{v \in V_i} \phi(v)
\end{equation}
where $\phi(\cdot)$ maps each concept $v$ to its processed surface form through a sequence of linguistic post-processing steps designed to preserve semantic coherence and ensure linguistic fluency. These include \textit{Temporal Expression Reconstruction}, where date and time expressions fragmented during AMR parsing are converted into natural language format, such as transforming "month 7 year 2025" into "July 2025"; \textit{Redundancy Removal}, which eliminates consecutive duplicate concepts to reduce repetition while maintaining semantic diversity; and \textit{Surface Realization}, which restores the processed concepts to their original textual forms in the raw document to mitigate potential distortions introduced by the AMR parsing process. This compressed form serves as the final input context, preserving the essential semantic signals while substantially reducing the original context length.

\section{Experiments}
\label{sec:experiments}

\subsection{Datasets and Implementation Details}
\label{sec:datasets}

We conduct comprehensive evaluations on two widely-adopted open-domain question-answering datasets that provide long-context supporting documents for RAG-based inference: \textbf{PopQA}~\cite{mallen-etal-2023-trust} and \textbf{EntityQuestions}~\cite{sciavolino-etal-2021-simple}. For comprehensive evaluation, we use Contriever~\cite{izacard2022unsupervised} as the retriever for PopQA and BM25~\cite{robertson2009probabilistic} for EntityQuestions, with retriever optimization beyond the scope of this work. Both datasets are equipped with ground-truth annotations indicating whether each supporting document contains the correct answer, denoted by the boolean indicator "\textit{hasanswer}". To align the problem formulation in Eq.~\ref{eq_formulation}, we retain only documents where "\textit{hasanswer" = True}, ensuring that performance variations stem from compression effectiveness rather than irrelevant document interference. For each query $Q$, let $K$ denote the number of answer-containing documents in the filtered $D$. The statistical characteristics of the curated $\langle Q, A, D \rangle$ triplets are summarized as follows:

\begin{table}[htbp]
\centering
\caption{Statistical results of the amount of screened-out $\langle Q, A, D \rangle$ pairs from the datasets.}
\begin{adjustbox}{width=1\linewidth}
\begin{tabular}{c|c|c|c|c|c|c|c|c|c|c}
\hline
$K$= & 1 & 2 & 3 & 4 & 5 & 6 & 7 & 8 & 9 & 10 \\ \hline \hline
PopQA & 280 & 298 & 174 & 172 & 160 & 153 & 149 & 155 & 135 & 125 \\ \hline
EQ & 489 & 572 & 373 & 295 & 239 & 199 & 179 & 169 & 130 & 113 \\
\hline
\end{tabular}
\label{tab:datasets}
\end{adjustbox}
\end{table}

To mitigate reliance on parametric knowledge in LLM inference, we employ a structured prompting that prioritizes externally provided evidence over internal memory. We adopt the instruction as follows: "[\textit{Refer to the following facts to answer the question. Facts: $C'$. Question: $Q$}]". Given that prompt intensity significantly influences inference behavior~\cite{wu2024clasheval}, we frame the supporting concepts $C'$ as "facts" to establish a constrained knowledge boundary that minimizes interference from potentially conflicting parametric knowledge.

\subsection{Baseline Methods}
\label{sec:baselines}

Our baseline evaluation examines two key dimensions: (1) diverse backbone LLM architectures, and (2) alternative context compression techniques. For backbone LLMs, we select mainstream publicly available LLMs, including GPT-Neo (1.3b and 2.7b)~\cite{gpt-neo}, OPT (1.3b and 2.7b)~\cite{zhang2022opt}, BLOOM LM (560m, 7b1)~\cite{le2022bloom}, LLaMA-2-chat (13b)~\cite{touvron2023llama}, Llama-3.1-Instruct (8b)~\cite{llama3}, DeepSeek-V2-Lite (16b)~\cite{deepseekv2}, and Qwen3 (32b)~\cite{qwen3technicalreport}. The combination of backbone LLMs with contexts in raw supporting documents constitutes the \textit{Vanilla} baseline.

For context compression, we implement five representative approaches that span different paradigms. We categorize these methods into three groups to answer the following questions: \textbf{Q1:} {Can simple frequency-based measures suffice for identifying informative content?} (\textit{Statistical Method}). \textbf{Q2:} {Can LLMs perform compression effectively through prompt-based reasoning?} (\textit{LLMs-driven Methods}). \textbf{Q3:} {Can dedicated context compression models be more targeted and effective?} (\textit{Compression-specific Methods}). 

The baselines corresponding to the above questions are as follows: (1) \textit{Statistical Method}: TF-IDF, the statistical entropy-inspired method that identifies salient terms using frequency–inverse document frequency weighting to highlight informative concepts. (2) \textit{LLMs-driven Methods}: prompt-based keyword extraction and summarization that leverage LLaMA-3.1-8B-Instruct with prompts as Prompt~\ref{prompt_keywords} and Prompt~\ref{prompt_summary} to generate keywords and summarizations. (3) \textit{Compression-specific Methods}: Selective Context (SelCon)~\cite{li-etal-2023-compressing} that employs trained models to identify relevant spans, and LLMLingua~\cite{jiang-etal-2023-llmlingua} uses budget-constrained token selection for optimal compression. These baselines evaluate if compressed contexts can preserve essential information while reducing computational overhead.

\subsection{Evaluation Metrics}
\label{sec:evaluation}

We employ three metrics to evaluate performance: accuracy (\texttt{Acc}), Area Under the Curve (\texttt{AUC}), and standard deviation ($\sigma$) of \texttt{AUC} as an auxiliary metric. The \texttt{Acc} follows the exact match protocol of~\citet{mallen-etal-2023-trust}, measuring if any generated answer exactly matches any gold-standard $a_j \in A$ for a given query $Q$. The $\sigma$ assesses the stability of compressed methods across different backbone LLMs.

The \texttt{AUC} provides a comprehensive assessment across varying $K$. Specifically, \texttt{AUC} computes the area under the \texttt{Acc} curve against $K$. Higher \texttt{AUC} indicates superior overall performance across the corresponding intervals. Given our focus on long-context compression, we partition the \texttt{AUC} calculation into two intervals for the values of $K$: a standard interval $I_s=[1,10]$ that captures general performance trends and a long-context interval $I_l=[6,10]$ that highlights performance under long context. This decomposition provides clear insights into both typical and challenging scenarios.

\section{Results and Analysis}

\subsection{Overall Performance}

\begin{table*}[!h]
\centering
\caption{The quantitative results of \texttt{AUC} $\uparrow$ for the PopQA dataset, where the full name order of the LLMs is: GPT-Neo-1.3B, GPT-Neo-2.7B, OPT-1.3b, OPT-2.7b, Bloom-560m, Bloom-7b1, Llama-2-chat-13b, Llama-3.1-8B-Instruct, DeepSeek-V2-Lite, Qwen3-32B. The standard division is as $\sigma \downarrow$. The best results are in \textbf{bold}, and the second-best results are in \uline{underlined}. The \textcolor{applegreen}{increased} and \textcolor{downred}{decreased} $\Delta$ are marked differently.}
\begin{adjustbox}{width=0.85\linewidth}
\begin{tabular}{ccccccccccccc}
\hline
$D$                  & $K$          & G-1.3 & G-2.7 & O-1.3 & O-2.7 & b-560 & b-7b1 & L-13 & L3.1-8 & DS-V2 & Q3-32 & $\sigma \downarrow$ \\ \hline \hline
\multirow{2}{*}{\rotatebox{0}{\footnotesize{Vanilla}}} & {$I_s$} & 553.32 & 550.79 & 585.12 & 596.31 & \underline{575.04} & 664.92 & 583.57 & 701.36 & 575.00 & 251.99 &  119.63  \\ 
                        
                         & {$I_l$} & 262.07 & 252.04 & 278.86 & 282.63 & \underline{284.04} & \underline{318.37} & 293.42 & 337.14 & 303.30 & 101.33 & 64.77  \\ \hline
\multirow{2}{*}{\rotatebox{0}{\footnotesize{TF-IDF}}}     & {$I_s$} & 354.04 & 508.48 & 486.22 & 523.84 & 417.67 & 608.85 & 623.00 & 650.98 & 179.28 & 210.62 & 165.39  \\ 
                         
                         & {$I_l$} & 169.82 & 251.12 & 244.02 & 269.09 & 217.52 & 307.70 & 311.47 & 316.14 & 106.47 & 113.97 & 78.00   \\ \hline
\multirow{2}{*}{\rotatebox{0}{\footnotesize{Keywords}}}     & {$I_s$} &  423.52 & 449.40 & 532.66 & 547.01 & 497.93 & 588.64 & 552.55 & 606.34 & 295.62 & 271.88 & 116.40   \\ 
                         
                         & {$I_l$} & 193.41 & 211.08 & 264.65 & 274.44 & 252.10 & 294.34 & 278.92 & 302.44 & 173.88 & 141.73 & 55.10   \\ \hline
\multirow{2}{*}{\rotatebox{0}{\footnotesize{Summary}}}      & {$I_s$} &  433.24 & 459.55 & 540.52 & 504.34 & 527.49 & 577.91 & 482.79 & 551.42 & 491.56 & 285.17 &  \textbf{82.74}   \\ 
                         
                         & {$I_l$} &  206.04 & 223.84 & 267.55 & 242.91 & 268.18 & 294.93 & 252.27 & 270.41 & 269.50 & 138.74 & \underline{44.81}   \\ \hline
\multirow{2}{*}{\rotatebox{0}{\footnotesize{SelCon}}}      & {$I_s$} &  453.31 & 490.44 & 580.08 & 581.62 & 443.08 & 634.40 & 637.20 & 717.74 & 557.43 & 293.10 & 121.98   \\ 
                         
                         & {$I_l$} &  209.18 & 228.22 & 286.68 & 284.62 & 216.25 & 307.80 & 309.70 & 339.02 & 295.48 & 156.93 & 57.34   \\ \hline
\multirow{2}{*}{\rotatebox{0}{\footnotesize{Lingua}}}      & {$I_s$} &  \underline{554.94} & \underline{553.15} & \underline{607.40} & \underline{617.07} & 567.67 & \underline{665.73} & \underline{645.21} & \underline{743.76} & \underline{643.01} & \underline{325.39} & 110.21   \\ 
                         
                         & {$I_l$} &  \underline{263.89} & \underline{258.09} & \underline{292.36} & \underline{286.70} & 280.85 & 317.55 & \underline{312.28} & \underline{346.24} & \textbf{318.18} & \underline{163.83} & 50.08    \\ \hline
\multirow{2}{*}{\rotatebox{0}{\footnotesize{\textbf{Ours}}}}    & \cellcolor{intercolor_1}{$I_s$} & \cellcolor{intercolor_1}\textbf{600.62} & \cellcolor{intercolor_1}\textbf{611.43} & \cellcolor{intercolor_1}\textbf{625.14} & \cellcolor{intercolor_1}\textbf{648.91} & \cellcolor{intercolor_1}\textbf{587.98} & \cellcolor{intercolor_1}\textbf{677.77} & \cellcolor{intercolor_1}\textbf{678.51} & \cellcolor{intercolor_1}\textbf{756.44} & \cellcolor{intercolor_1}\textbf{648.90} & \cellcolor{intercolor_1}\textbf{356.55} & \cellcolor{intercolor_1}\underline{104.32} \\ 
                         
                         & \cellcolor{intercolor_2}{$I_l$} & \cellcolor{intercolor_2}\textbf{283.54} & \cellcolor{intercolor_2}\textbf{296.09} & \cellcolor{intercolor_2}\textbf{298.73} & \cellcolor{intercolor_2}\textbf{308.92} & \cellcolor{intercolor_2}\textbf{292.74} & \cellcolor{intercolor_2}\textbf{332.16} & \cellcolor{intercolor_2}\textbf{326.67} & \cellcolor{intercolor_2}\textbf{357.74} & \cellcolor{intercolor_2}\underline{318.06} & \cellcolor{intercolor_2}\textbf{191.09} &  \cellcolor{intercolor_2}\textbf{44.33} \\ \hline \hline
\multirow{2}{*}{\rotatebox{0}{\footnotesize{$\Delta$}}}    & {$I_s$} & \textcolor{applegreen}{+47.30} & \textcolor{applegreen}{+60.64}  & \textcolor{applegreen}{+40.02} & \textcolor{applegreen}{+52.60} & \textcolor{applegreen}{+12.94} & \textcolor{applegreen}{+12.85} & \textcolor{applegreen}{+94.94} & \textcolor{applegreen}{+55.08} & \textcolor{applegreen}{+73.90} & \textcolor{applegreen}{+104.56} & 30.32 \\ 
                         
                         & {$I_l$} & \textcolor{applegreen}{+21.47} & \textcolor{applegreen}{+44.05} & \textcolor{applegreen}{+19.87} & \textcolor{applegreen}{+26.29} & \textcolor{applegreen}{+8.70} & \textcolor{applegreen}{+13.79} & \textcolor{applegreen}{+33.25} & \textcolor{applegreen}{+20.60} & \textcolor{applegreen}{+14.76} & \textcolor{applegreen}{+89.76} & 23.57 \\ \hline
\end{tabular}
\end{adjustbox}
\label{tab:PopQA}
\end{table*}

\begin{table*}[!h]
\centering
\caption{The \texttt{AUC} $\uparrow$ results for the EntityQuestions dataset. The symbol definitions are same as Table~\ref{tab:PopQA}.}
\begin{adjustbox}{width=0.85\linewidth}
\begin{tabular}{ccccccccccccc}
\hline
$D$                  & $K$          & G-1.3 & G-2.7 & O-1.3 & O-2.7 & b-560 & b-7b1 & L-13 & L3.1-8 & DS-V2 & Q3-32 & $\sigma \downarrow$ \\ \hline \hline
\multirow{2}{*}{\rotatebox{0}{\footnotesize{Vanilla}}} & {$I_s$} & \textbf{550.08} & \underline{608.54} & \underline{618.05} & \textbf{677.63} & \textbf{511.98} & \textbf{705.35} & 657.06 & 743.99 & 572.72 & 235.42 & 142.98  \\ 
                        
                         & {$I_l$} & \textbf{259.35} & \underline{283.86} & \underline{284.91} & \textbf{318.26} & \textbf{236.82} & \textbf{329.58} & 296.63 & 338.60 & \textbf{313.36} & 87.65 & 72.88    \\ \hline
\multirow{2}{*}{\rotatebox{0}{\footnotesize{TF-IDF}}}     & {$I_s$} & 302.59 & 459.72 & 419.50 & 517.23 & 314.45 & 552.43 & 666.08 & 627.44 & 180.75 & 235.64 & 165.91    \\ 
                         
                         & {$I_l$} & 146.52 & 239.16 & 188.60 & 259.91 & 155.99 & 273.13 & \underline{323.23} & 276.02 & 107.46 & 112.64 & 75.92    \\ \hline
\multirow{2}{*}{\rotatebox{0}{\footnotesize{Keywords}}}     & {$I_s$} &  358.34 & 458.67 & 495.48 & 545.41 & 392.71 & 572.18 & 614.18 & 674.23 & 284.15 & 287.12 & 135.78    \\ 
                         
                         & {$I_l$} & 171.09 & 229.08 & 245.89 & 276.19 & 190.40 & 282.74 & 310.78 & 323.42 & 175.65 & 128.99 & 65.40    \\ \hline
\multirow{2}{*}{\rotatebox{0}{\footnotesize{Summary}}}      & {$I_s$} &  336.92 & 366.90 & 450.84 & 437.40 & 396.18 & 498.25 & 435.01 & 511.30 & 448.16 & 210.08 & \textbf{88.12}    \\ 
                         
                         & {$I_l$} &  161.38 & 180.04 & 221.94 & 202.50 & 196.11 & 254.38 & 209.77 & 242.42 & 247.76 & 77.62 & \textbf{52.17}    \\ \hline
\multirow{2}{*}{\rotatebox{0}{\footnotesize{SelCon}}}      & {$I_s$} &  278.08 & 329.18 & 359.08 & 391.45 & 251.39 & 401.26 & 531.96 & 545.13 & 395.29 & 226.52 & \underline{107.42}    \\ 
                         
                         & {$I_l$} &  136.32 & 163.02 & 177.21 & 187.91 & 137.72 & 195.78 & 268.26 & 259.44 & 208.08 & 103.98 & \underline{52.52}    \\ \hline
\multirow{2}{*}{\rotatebox{0}{\footnotesize{Lingua}}}      & {$I_s$} &  541.93 & 598.45 & 592.69 & 644.01 & \underline{496.46} & 670.92 & \underline{698.64} & \underline{792.93} & \underline{648.58} & \underline{374.74} & 115.86    \\ 
                         
                         & {$I_l$} &  244.38 & 275.40 & 274.64 & 283.11 & 223.05 & 308.36 & 322.57 & \underline{357.82} & \underline{307.12} & \underline{152.43} & 57.73  \\ \hline
\multirow{2}{*}{\rotatebox{0}{\footnotesize{\textbf{Ours}}}}    & \cellcolor{intercolor_1}{$I_s$} & \cellcolor{intercolor_1}\underline{546.46} & \cellcolor{intercolor_1}\textbf{627.41} & \cellcolor{intercolor_1}\textbf{632.79} & \cellcolor{intercolor_1}\underline{662.16} & \cellcolor{intercolor_1}{494.45} & \cellcolor{intercolor_1}\underline{688.73} & \cellcolor{intercolor_1}\textbf{738.82} & \cellcolor{intercolor_1}\textbf{813.86} & \cellcolor{intercolor_1}\textbf{652.14} & \cellcolor{intercolor_1}\textbf{406.00} & \cellcolor{intercolor_1}118.33 \\ 
                         
                         & \cellcolor{intercolor_2}{$I_l$} & \cellcolor{intercolor_2}\underline{248.82} & \cellcolor{intercolor_2}\textbf{294.48} & \cellcolor{intercolor_2}\textbf{298.31} & \cellcolor{intercolor_2}\underline{295.18} & \cellcolor{intercolor_2}\underline{229.06} & \cellcolor{intercolor_2}\underline{323.26} & \cellcolor{intercolor_2}\textbf{343.58} & \cellcolor{intercolor_2}\textbf{371.30} & \cellcolor{intercolor_2}{307.05} & \cellcolor{intercolor_2}\textbf{181.50} & \cellcolor{intercolor_2}55.95 \\ \hline \hline
\multirow{2}{*}{\rotatebox{0}{\footnotesize{$\Delta$}}}    & {$I_s$} & \textcolor{downred}{-3.62} & \textcolor{applegreen}{+18.87}  & \textcolor{applegreen}{+14.74} & \textcolor{downred}{-15.47} & \textcolor{downred}{-17.53} & \textcolor{downred}{-16.62} & \textcolor{applegreen}{+81.76} & \textcolor{applegreen}{+69.87} & \textcolor{applegreen}{+79.42} & \textcolor{applegreen}{+170.58} & 61.27 \\ 
                         
                         & {$I_l$} & \textcolor{downred}{-10.53} & \textcolor{applegreen}{+10.62} & \textcolor{applegreen}{+13.40} & \textcolor{downred}{-23.08} & \textcolor{downred}{-7.76} & \textcolor{downred}{-6.32} & \textcolor{applegreen}{+46.95} & \textcolor{applegreen}{+32.70} & \textcolor{downred}{-6.31} & \textcolor{applegreen}{+93.85} & 35.11 \\ \hline
\end{tabular}
\end{adjustbox}
\label{tab:EQQA}
\end{table*}

The \texttt{AUC} results in $I_s$ interval in Table~\ref{tab:PopQA} and Table~\ref{tab:EQQA} present the overall performance comparison across both datasets. The full results in \texttt{Acc} are in Table~\ref{tab:aca_PopQA} and~\ref{tab:aca_EQQA} respectively. In the PopQA dataset, the proposed method achieves substantial gains compared to the vanilla baseline. The most notable improvements occur in larger models like Qwen3-32B, Llama-2-chat-13b, and DeepSeek-V2-Lite. In contrast, smaller models like Bloom-560m/7b1 show relatively modest improvements. On the EntityQuestions dataset, the results exhibit similar trends with some variations. The proposed method achieves the best or second-best performance across most configurations, with particularly strong results on larger models like Qwen3-32B. However, we observe slight performance degradation compared to vanilla on smaller models like GPT-Neo-1.3B and Bloom-560m/7b1. Considering the previous observation, this phenomenon indicates that compact LLMs may benefit from more contextual information that retains rich linguistic elements to reconstruct scenarios rather than aggressive compression. This suggests a trade-off between compression ratio and model capacity that warrants consideration in practical deployments. In addition, our method achieves a competitive $\sigma$ across diverse backbone LLMs, indicating it preserves universally shared semantic cores rather than model-specific preferences, forming a robust semantic compression that maintains coherent reasoning chains across different architectures.

Compared to compression baselines, our method demonstrates substantial advantages across different paradigms. Against the statistical TF-IDF approach, we achieve overwhelming superiority on both datasets, outperforming all backbone LLMs. Although TF-IDF outperforms the vanilla setting on certain backbone models, this improvement is not consistent when examined across different architectures, as indicated by the unstable results with the highest $\sigma$. Its performance depends on surface-level lexical patterns, which may occasionally align with answer-bearing spans in simple contexts. However, TF-IDF lacks semantic structure awareness and does not model how LLMs reconstruct contextual meaning. As a result, it may either discard essential cues or retain redundant tokens that vary across models. The fluctuating performance across backbones indicates answers of \textbf{Q1} that frequency-based signals are insufficient for reliably identifying informative content.

The LLM-driven baselines, Keywords and Summary, show limited performance in most settings. Unlike statistical measures, these baselines depend on generative rewriting, which makes them sensitive to semantic integrity and prompts. These factors lead to unreliable results across different backbones. In addition, the generative paradigm can introduce hallucinations into the rewritten content, further increasing the uncertainty of the compressed context. A notable trend is that the summary-based compression achieves the lowest $\sigma$. The reason is summary-compressed context is remain natural language, forming a continuous representation showing lower sensitivity to surface-level changes. In contrast, the discrete keywords-based compression shows notable performance swings. These observations answer \textbf{Q2} by showing that LLM-driven baselines are not a reliable choice due to the uncertainty in inference.

Compared with the SelCon baseline, our method achieves higher \texttt{AUC} across configurations. We hypothesize that this gap stems from fundamental differences in our approaches: while both methods utilize information theory, SelCon operates at the phrase/sentence level through token-based self-information aggregation for content filtering, whereas our method uses AMR's structured semantic representation to compute concept-level entropy based on semantic roles and connections in comprehensive contexts. The AMR-based entropy better preserves the conceptual coherence for complex reasoning, as it captures semantic structures and dependencies that are crucial for maintaining clear inferential chains for reconstructing scenarios.

LLMLingua represents competitive baseline as a token-level compression technique. The advantage of our method relative to LLMLingua comes from the complementary strengths of semantic-level versus token-level compression: while LLMLingua selects tokens through iterative perplexity-based filtering and budget control, our AMR-based approach identifies coherent concept units that match the information structure. Both methods preserve essential information, but our semantic abstraction excels when maintaining conceptual relationships matters more than surface-level linguistic continuity. Moreover, our method enhances the interpretability and readability by preserving complete conceptual units as atomic elements and maintaining lexical integrity, whereas token-level compression can fragment words that disrupt local linguistic structures. This property facilitates human understanding and debugging. Compared with other baselines, both SelCon and LLMLingua achieve competitive \texttt{AUC} and $\sigma$, addressing \textbf{Q3} on the necessity of dedicated context compression methods.

\subsection{Performance on Long Contexts}

To further validate our method and highlight its characteristics, we analyze performance in the long-context interval $I_l$ in Table~\ref{tab:PopQA} and Table~\ref{tab:EQQA}, emphasizing behaviors that emerge specifically under long-context conditions. The proposed method achieves the competitive performance that keeps the same trend as in the $I_s$, but the gains are reduced. The reduction is expected since the $I_l$ interval typically encompasses longer contexts or higher complexity scenarios, where the marginal benefit of improvements tends to diminish. However, a notable phenomenon is that $\sigma$ is significantly lower for this interval, which contains longer but more concentrated concepts compared with the massive but dispersed interval, indicating the benefit of macro-level semantic constraints in capturing informative concepts within complex contexts in specific scenarios. Moreover, the low $\sigma$ of $\Delta$ indicates consistent performance variance across backbones.

\subsection{Compression Efficiency}

\begin{figure}[h]
\centering
\includegraphics[width=1\linewidth]{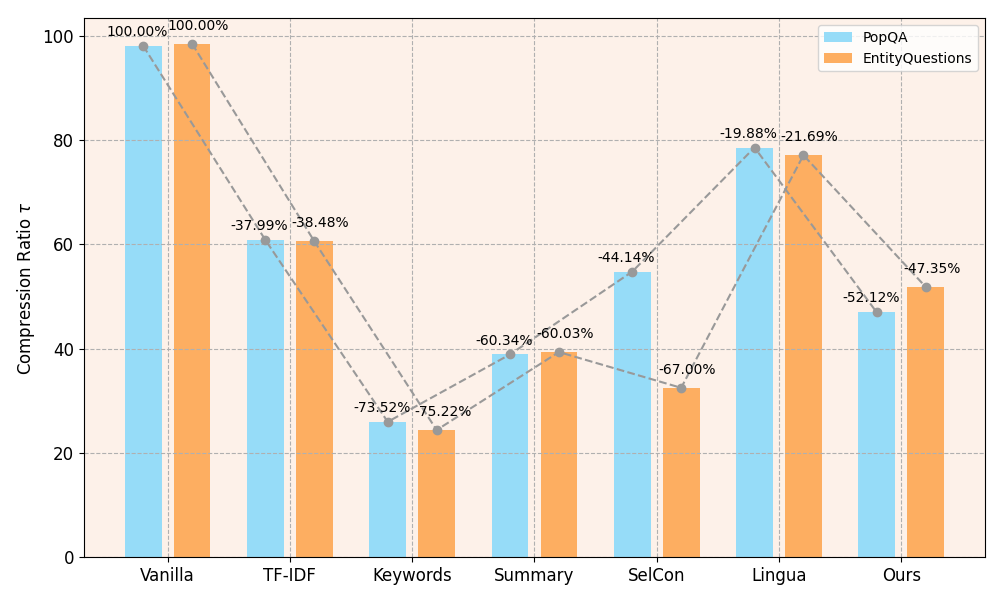}
\caption{Comparison of token-level compression ratios across different context compression methods.}
\label{fig:compression_ratio}
\end{figure}

We examine the compression efficiency in terms of token-level reduction ($\tau$) and inference latency (ms per instance). As shown in Figure~\ref{fig:compression_ratio}, our method reduces the length to about $50\%$ of the vanilla on average, while keeping the \texttt{Acc} stable in both datasets. Baselines such as Keywords and Summary yield lower token counts, but they often remove meaningful factual cues, leading to performance drops. In contrast, operating at the concept level through AMR allows the compressed context to retain the core semantic units needed for reasoning, rather than relying on surface lexical signals.

\begin{table}[h]
\centering
\caption{Inference time comparison (ms per instance)}
\begin{adjustbox}{max width=\linewidth}
\begin{tabular}{lccccccc}
\toprule
\textbf{LLMs} & \textbf{Vanilla} & \textbf{TF-IDF} & \textbf{Keywords} & \textbf{Summary} & \textbf{SelCon} & \textbf{Lingua} & \textbf{Ours} \\
\hline \hline
\multicolumn{8}{c}{PopQA} 
\\ \hline
G-1.3 & 402.89  & 468.01 & \textbf{366.23}   & 410.52  & 470.27 & 429.32 & \underline{380.38} \\
G-2.7 & 672.12  & 622.81 & \textbf{548.13}   & 578.64  & 634.32 & 640.69 & \underline{548.51} \\
O-1.3 & 322.68  & 314.18 & \textbf{281.84}   & 316.45  & \underline{305.92} & 356.40 & 306.23 \\
O-2.7 & 517.73  & 499.23 & \underline{484.59}   & 487.71  & 524.98 & 526.04 & \textbf{461.01} \\
b-560 & 261.43  & 265.57 & \textbf{235.13}   & \underline{237.32}  & 275.10 & 274.51 & 249.55 \\
b-7b1 & 1130.13 & 1152.86 & \textbf{1006.23}   & \underline{1006.60}  &1150.33 &1139.21 & 1058.83 \\
L-13  & 1886.29 & 1405.44 & \textbf{1329.71}   & \underline{1364.58}  &1476.61 &1507.88 & 1409.22 \\
L3.1-8 & 1032.17 & 1091.39 & \underline{688.09}   & \textbf{644.89}  &1109.62 &1089.12 & 888.58 \\
DS-V2 & 1233.80 & 166.51 & \textbf{150.13}   & 165.69  & 293.25 & 171.14 & \underline{164.05} \\
Q3-32 & 5283.06 & 5029.57 & \textbf{4795.76}   & 4879.41  & 5094.38 & 5040.01 & \underline{4783.34} \\
\hline \hline
\multicolumn{8}{c}{EntityQuestions} 
\\ \hline
G-1.3 & 605.82 & 587.49 & \textbf{546.79} & \underline{547.65} & 724.10 & 761.94 & 585.63 \\
G-2.7 & 866.79 & 811.66 & \underline{749.83} & \textbf{746.46} & 867.93 & 932.25 & 779.43 \\
O-1.3 & 528.14 & \textbf{486.72} & \underline{481.75} & 496.73 & 557.98 & 648.45 & 499.03 \\
O-2.7 & 703.28 & 684.91 & \textbf{647.43} & \underline{671.47} & 761.57 & 827.20 & 702.26 \\
b-560 & 445.16 & 468.14 & \underline{421.71} & \textbf{416.70} & 527.12 & 582.26 & 439.89 \\
b-7b1 & 1319.82 & 1338.23 & 1196.88 & \textbf{1176.98} & \underline{1190.92} & 1456.20 & 1279.33 \\
L-13 & 1805.86 & 1786.85 & \underline{1672.17} & 1743.26 & 1717.31 & 1881.69 & \textbf{1590.65} \\
L3.1-8 & 1233.70 & 1282.96 & \underline{871.17} & \textbf{836.18} & 1016.64 & 1398.92 & 1083.90 \\
DS-V2 & 358.03 & \textbf{326.23} & 333.82 & \underline{326.80} & 431.81 & 444.87 & 330.10 \\
Q3-32 & 5239.69 & 5313.41 & \textbf{4996.61} & \underline{5012.55} & 5120.52 & 5409.85 & 5168.47 \\
\hline
\end{tabular}
\label{tab:inference_time}
\end{adjustbox}
\end{table}

The reduction in context length leads directly to faster inference, and the latency decreases in line with the length reduction. Table~\ref{tab:inference_time} shows that the proposed method lowers the average inference time compared to the vanilla setting. Baselines reducing latency via token pruning may fragment expressions and weaken local coherence, especially in long contexts. By retaining intact conceptual units, our compressed contexts remain stable for reasoning, enabling both shorter inference time and reliable answering, even under high compression.

\section{Conclusion}

This paper presents a compression method for context engineering that leverages conceptual information entropy of AMR to identify semantically crucial concepts. Our method shows improvements over baselines while achieving substantial compression ratios. The experiments demonstrate that AMR-based semantic analysis guides context compression effectively. The integration of structured linguistic representation with information-theoretic concept selection offers a paradigm to balance information retention with computational efficiency.

Future research includes extending our approach to multi-modal contexts, modeling cross-document concept relationships, and exploring adaptive compression strategies based on query complexity. Incorporating other stable linguistic representations is also a valuable direction to improve the efficiency and effectiveness in context engineering.

\section*{Limitations}

Although the proposed method shows clear gains in long-context settings, some limitations remain. First, the current approach relies on the stability of AMR parsers, and the performance may decline when the parser produces incomplete or noisy graphs. The parsing processing is based on the sentence-level graph, so complex document-level structures are easily ignored. These dependency introduces upper bounds on covered conceptual information in compression. Developing reliable AMR parsers is a continuously valuable direction.

Second, the current setup evaluates compression under a controlled testing environment where answer-containing documents are considered. This design isolates the effect of compression but does not fully reflect real-world retrieval pipelines, where irrelevant or conflicting documents are common. Experimenting with the setting in a full retrieval stack and examining different retrievers' influence will be conducted in future work.

Finally, computing AMR graphs and entropy scores introduces extra cost during preprocessing. Although this cost occurs offline, it may restrict the method in latency-sensitive systems or in large-scale applications where many documents must be processed. A crucial future work is exploring high-efficiency solutions for these stages.

\section*{Acknowledgments}

This work is supported by the 2025 UniSQ Academic Affairs Collaboration Grants.

\bibliography{custom}

\clearpage
\newpage

\appendix

\setcounter{figure}{0}
\setcounter{table}{0}
\setcounter{mytcolorbox}{0}
\renewcommand{\thefigure}{A\arabic{figure}}
\renewcommand{\thetable}{A\arabic{table}}
\renewcommand{\themytcolorbox}{A\arabic{mytcolorbox}}

\section{Prompts for Baselines}
\label{sec:prompt}

Following the instruction-tuning framework of~\citet{alpaca}, we design prompt templates for keyword extraction and summarization baselines, as detailed in Prompt~\ref{prompt_keywords} and Prompt~\ref{prompt_summary}.

\begin{tcolorbox}[
    colframe=gray,  
    fonttitle=\bfseries, 
    title={Prompt A1: Keywords Extraction}, 
    before upper=\refstepcounter{mytcolorbox}
]
\label{prompt_keywords}
{[INST] <<SYS>> \\ \texttt{Extract a few keywords from the following content.}}\\ 
\text{<</SYS>>} \\
{Prompt = """Below is an instruction that describes a task, paired with an input that provides content.\\     \#\#\# Instruction: \{""" + \texttt{Instruction} + """\}\\    \#\#\# Input: \{""" + $D$ + """\}\\     \#\#\# Response: """} \\
\text{[/INST]} 
\end{tcolorbox}

\begin{tcolorbox}[
    colframe=gray, 
    fonttitle=\bfseries, 
    title={Prompt A2: Summary Generation}, 
    before upper=\refstepcounter{mytcolorbox}
]
\label{prompt_summary}
{[INST] <<SYS>> \\ \texttt{Generate a short summary of the following content.}}\\ 
\text{<</SYS>>} \\
{Prompt = """Below is an instruction that describes a task, paired with an input that provides content.\\     \#\#\# Instruction: \{""" + \texttt{Instruction} + """\}\\    \#\#\# Input: \{""" + $D$ + """\}\\     \#\#\# Response: """} \\
\text{[/INST]} 
\end{tcolorbox}

\section{Accuracy Details}
\label{sec:acc}

\begin{table*}[htbp]
\tiny
\centering
\caption{Accuracy (\texttt{Acc} $\uparrow$) comparison on the PopQA dataset. The best results for each LLM with setting $K$ are in \textbf{bold}, and the next best results are in \uline{underlined}. $\Delta$ here represents the difference between Ours and Vanilla, and the \textcolor{applegreen}{increased} and \textcolor{downred}{decreased} $\Delta$ are marked differently. The best results for each of $K$ are \textcolor{bestcolor}{marked}.}
\label{tab:aca_PopQA}
\renewcommand{\arraystretch}{0.25}
\begin{adjustbox}{width=0.82\linewidth}
\begin{tabular}{c|c|*{10}{c}}
\hline
LLMs & \diagbox{$C^{'}$}{$K$} & 1 & 2 & 3 & 4 & 5 & 6 & 7 & 8 & 9 & 10 \\
\hline
\multirowcell{30}{\rotatebox{90}{GPT-Neo-1.3B}} & Vanilla & 48.57 & \uline{64.77} & 54.60 & 54.07 & 63.13 & 60.78 & 62.42 & 63.23 & \textbf{69.63} & 72.80 \\
 & TF-IDF & 22.86 & 32.89 & 38.51 & 38.95 & 42.50 & 39.87 & 42.28 & 38.71 & 43.70 & 50.40 \\
 & Keywords & 30.00 & 41.61 & 43.68 & 50.58 & 53.75 & 50.98 & 48.32 & 47.10 & 43.70 & 57.60 \\
 & Summary & 23.93 & 41.61 & 47.13 & 48.84 & 52.50 & 50.33 & 46.98 & 48.39 & 56.30 & 58.40 \\
 & SelCon & 39.29 & 54.03 & 42.53 & 48.26 & 51.88 & 55.56 & 46.98 & 49.03 & 52.59 & 65.60 \\
 & LLMLingua & \textbf{53.93} & 53.69 & \uline{56.90} & \uline{58.72} & \uline{64.38} & 60.78 & 63.76 & 65.81 & 65.93 & 76.00 \\
 & Ours & \cellcolor{markcolor}\textbf{53.93} & \cellcolor{markcolor}\textbf{68.45} & \cellcolor{markcolor}\textbf{57.47} & \cellcolor{markcolor}\textbf{59.88} & \cellcolor{markcolor}\textbf{70.00} & \cellcolor{markcolor}\uline{68.63} & \cellcolor{markcolor}\textbf{69.13} & \cellcolor{markcolor}\textbf{71.61} & \cellcolor{markcolor}\uline{68.89} & \cellcolor{markcolor}\textbf{79.20} \\
 & $\alpha=0.01$ & 19.64 & 32.21 & 42.53 & 48.84 & 53.13 & 54.90 & 65.10 & 67.10 & 62.96 & 76.80 \\
 & $\alpha=0.05$ & 28.57 & 41.28 & 48.28 & 56.98 & 58.75 & 60.78 & 67.11 & \uline{68.39} & 63.70 & \uline{77.60} \\
 & $\alpha=0.1$ & 28.57 & 41.28 & 48.28 & 56.98 & 58.75 & 60.78 & 67.11 & \uline{68.39} & 63.70 & \uline{77.60} \\
 & $\alpha=0.5$ & 35.00 & 51.01 & \uline{56.90} & 58.14 & 61.88 & \textbf{69.28} & \textbf{69.13} & 62.58 & 65.19 & 76.80 \\
 & $\Delta$ & \textcolor{applegreen}{+5.36} & \textcolor{applegreen}{+3.68} & \textcolor{applegreen}{+2.87} & \textcolor{applegreen}{+5.81} & \textcolor{applegreen}{+6.87} & \textcolor{applegreen}{+7.85} & \textcolor{applegreen}{+6.71} & \textcolor{applegreen}{+8.38} & \textcolor{downred}{-0.74} & \textcolor{applegreen}{+6.40} \\
\hline
\multirowcell{30}{\rotatebox{90}{GPT-Neo-2.7B}} & Vanilla & \uline{51.07} & 69.46 & 59.77 & 52.91 & 59.38 & 63.40 & 59.73 & 57.42 & 65.19 & 76.00 \\
 & TF-IDF & 33.57 & 46.31 & 50.00 & 53.49 & 58.75 & 64.05 & 59.06 & 61.29 & 60.74 & 76.00 \\
 & Keywords & 30.71 & 43.00 & 48.28 & 51.16 & 54.38 & 52.29 & 56.38 & 47.10 & 51.85 & 59.20 \\
 & Summary & 22.86 & 39.93 & 50.00 & 50.00 & 56.25 & 56.21 & 52.35 & 57.42 & 55.56 & 60.80 \\
 & SelCon & 45.00 & 56.71 & 50.00 & 46.51 & 59.38 & 54.25 & 57.72 & 54.84 & 53.33 & 70.40 \\
 & LLMLingua & \textbf{52.14} & \uline{70.13} & 59.20 & 50.54 & 59.38 & 59.48 & 63.76 & 61.29 & 63.70 & 79.20 \\
 & Ours & \cellcolor{markcolor}\uline{51.07} & \cellcolor{markcolor}\textbf{70.45} & \cellcolor{markcolor}\uline{60.34} & \cellcolor{markcolor}\uline{61.63} & \cellcolor{markcolor}\textbf{64.38} & \cellcolor{markcolor}\textbf{66.01} & \cellcolor{markcolor}\textbf{75.17} & \cellcolor{markcolor}\uline{69.68} & \cellcolor{markcolor}\textbf{77.03} & \cellcolor{markcolor}\textbf{82.40} \\
 & $\alpha=0.01$ & 20.71 & 32.21 & 40.23 & 55.23 & 53.13 & 60.12 & 59.06 & 64.52 & 62.96 & 77.60 \\
 & $\alpha=0.05$ & 29.64 & 41.61 & 53.45 & 56.98 & 60.63 & 63.40 & 68.46 & 64.52 & 65.93 & 76.00 \\
 & $\alpha=0.1$ & 30.36 & 47.99 & 50.57 & 59.30 & 62.50 & 64.05 & \uline{72.48} & \textbf{71.61} & \uline{71.11} & \uline{81.60} \\
 & $\alpha=0.5$ & 40.36 & 53.02 & \textbf{61.49} & \textbf{64.53} & \textbf{64.38} & \textbf{66.01} & \uline{72.48} & 67.74 & 69.63 & \uline{81.60} \\
 & $\Delta$ & 0.00 & \textcolor{applegreen}{+0.99} & \textcolor{applegreen}{+0.57} & \textcolor{applegreen}{+8.72} & \textcolor{applegreen}{+5.00} & \textcolor{applegreen}{+2.61} & \textcolor{applegreen}{+15.44} & \textcolor{applegreen}{+12.26} & \textcolor{applegreen}{+11.84} & \textcolor{applegreen}{+6.40} \\
\hline
\multirowcell{30}{\rotatebox{90}{OPT-1.3b}} & Vanilla & 52.14 & 67.11 & \uline{63.22} & 57.56 & 61.25 & 62.09 & 69.13 & 71.62 & 66.67 & 80.80 \\
 & TF-IDF & 36.79 & 47.32 & 46.55 & 47.09 & 53.75 & 58.17 & 59.06 & 60.00 & 61.48 & 68.80 \\
 & Keywords & 31.79 & 45.30 & 55.17 & 55.23 & 64.38 & 64.05 & 65.10 & 68.39 & 60.74 & 76.80 \\
 & Summary & 26.79 & 47.65 & 56.90 & 59.30 & 65.00 & 61.44 & 65.10 & 67.74 & 65.19 & 77.60 \\
 & SelCon & 49.29 & 61.07 & 56.90 & 56.98 & 63.75 & 60.13 & 67.79 & 73.55 & \uline{74.07} & 82.40 \\
 & LLMLingua & \textbf{55.00} & \textbf{71.14} & 61.49 & 57.56 & 65.00 & 64.71 & \textbf{75.17} & 73.54 & 68.89 & \uline{84.80} \\
 & Ours & \cellcolor{markcolor}\uline{54.29} & \cellcolor{markcolor}\uline{69.13} & \cellcolor{markcolor}\textbf{68.39} & \cellcolor{markcolor}59.30 & \cellcolor{markcolor}\textbf{68.13} & \cellcolor{markcolor}\uline{68.62} & \cellcolor{markcolor}74.50 & \cellcolor{markcolor}74.19 & \cellcolor{markcolor}73.33 & \cellcolor{markcolor}\uline{84.80} \\
 & $\alpha=0.01$ & 23.57 & 34.56 & 44.25 & 48.84 & 58.13 & 60.13 & 69.80 & 73.55 & 70.37 & \uline{84.80} \\
 & $\alpha=0.05$ & 30.36 & 44.30 & 55.17 & \uline{59.88} & 60.00 & 62.09 & 73.83 & 73.55 & 70.37 & 82.40 \\
 & $\alpha=0.1$ & 32.86 & 46.98 & 60.92 & \textbf{62.79} & \uline{66.25} & \textbf{69.93} & 73.15 & \uline{75.48} & \textbf{75.56} & \textbf{86.40} \\
 & $\alpha=0.5$ & 42.14 & 57.72 & 60.34 & \uline{59.88} & 65.63 & 68.28 & \textbf{75.17} & \textbf{77.42} & 71.85 & 84.00 \\
 & $\Delta$ & \textcolor{applegreen}{+2.15} & \textcolor{applegreen}{+2.02} & \textcolor{applegreen}{+5.17} & \textcolor{applegreen}{+1.74} & \textcolor{applegreen}{+6.88} & \textcolor{applegreen}{+6.53} & \textcolor{applegreen}{+5.37} & \textcolor{applegreen}{+2.57} & \textcolor{applegreen}{+6.66} & \textcolor{applegreen}{+4.00} \\
\hline
\multirowcell{30}{\rotatebox{90}{OPT-2.7b}} & Vanilla & 49.64 & 66.78 & \uline{62.64} & 63.72 & 65.00 & 61.44 & 64.43 & 70.32 & 75.56 & \uline{83.20} \\
 & TF-IDF & 33.21 & 48.32 & 49.43 & 51.16 & 56.88 & 64.71 & 64.43 & 70.32 & 65.19 & 73.60 \\
 & Keywords & 35.36 & 43.96 & 58.05 & 58.14 & 65.00 & 59.48 & 66.44 & 70.97 & 68.89 & 76.80 \\
 & Summary & 29.64 & 49.33 & 54.60 & 55.23 & 60.00 & 54.90 & 60.40 & 65.16 & 56.30 & 67.20 \\
 & SelCon & 48.21 & 64.09 & 54.02 & 58.14 & 66.25 & 60.78 & 67.11 & 74.19 & 73.33 & 79.20 \\
 & LLMLingua & \textbf{55.71} & \textbf{73.15} & \uline{62.64} & 62.79 & \uline{71.25} & 65.36 & 63.09 & 77.42 & 71.11 & \textbf{84.80} \\
 & Ours & \cellcolor{markcolor}\uline{55.36} & \cellcolor{markcolor}\uline{70.13} & \cellcolor{markcolor}\textbf{64.37} & \cellcolor{markcolor}\textbf{70.35} & \cellcolor{markcolor}\textbf{72.50} & \cellcolor{markcolor}\uline{69.93} & \cellcolor{markcolor}\textbf{76.51} & \cellcolor{markcolor}\textbf{78.06} & \cellcolor{markcolor}\uline{77.78} & \cellcolor{markcolor}\uline{83.20} \\
 & $\alpha=0.01$ & 22.86 & 35.91 & 45.98 & 54.65 & 61.88 & 62.75 & 66.44 & 72.26 & 68.15 & \uline{83.20} \\
 & $\alpha=0.05$ & 33.57 & 45.30 & 59.77 & 61.05 & 66.25 & 66.67 & 73.15 & \textbf{78.06} & \uline{77.78} & 80.80 \\
 & $\alpha=0.1$ & 35.00 & 53.02 & 60.34 & 66.86 & 69.38 & 67.32 & \uline{75.17} & 76.13 & \textbf{79.26} & 80.00 \\
 & $\alpha=0.5$ & 45.36 & 63.42 & \uline{62.64} & \uline{68.02} & 66.25 & \textbf{73.20} & 73.83 & 71.61 & \uline{77.78} & 80.00 \\
 & $\Delta$ & \textcolor{applegreen}{+5.72} & \textcolor{applegreen}{+3.35} & \textcolor{applegreen}{+1.73} & \textcolor{applegreen}{+6.63} & \textcolor{applegreen}{+7.50} & \textcolor{applegreen}{+8.49} & \textcolor{applegreen}{+12.08} & \textcolor{applegreen}{+7.74} & \textcolor{applegreen}{+2.22} & 0.00 \\
\hline
\multirowcell{30}{\rotatebox{90}{Bloom-560m}} & Vanilla & 51.07 & 62.42 & 54.02 & \textbf{56.40} & 61.25 & \uline{62.75} & 66.44 & 72.90 & \textbf{73.33} & \uline{80.00} \\
 & TF-IDF & 27.14 & 34.90 & 36.78 & 43.60 & 48.75 & 45.10 & 57.72 & 52.26 & 52.59 & 64.80 \\
 & Keywords & 26.43 & 44.30 & \textbf{56.32} & 48.26 & 55.63 & 56.21 & 62.42 & 63.23 & 60.74 & 75.20 \\
 & Summary & 27.50 & 47.65 & 52.87 & 54.07 & \uline{61.88} & 58.17 & 67.11 & 70.97 & 62.22 & 77.60 \\
 & SelCon & 34.29 & 49.66 & 45.40 & 43.60 & 47.50 & 47.06 & 51.68 & 59.35 & 48.89 & 65.60 \\
 & LLMLingua & \textbf{53.57} & \uline{65.10} & 52.87 & 52.33 & 60.00 & 59.48 & 68.46 & \uline{74.84} & 67.41 & \textbf{80.80} \\
 & Ours & \cellcolor{markcolor}\uline{52.86} & \cellcolor{markcolor}\textbf{66.44} & \cellcolor{markcolor}\uline{55.74} & \cellcolor{markcolor}\uline{55.23} & \cellcolor{markcolor}59.38 & \cellcolor{markcolor}\textbf{64.05} & \cellcolor{markcolor}\textbf{71.81} & \cellcolor{markcolor}\textbf{76.77} & \cellcolor{markcolor}\textbf{73.33} & \cellcolor{markcolor}77.60 \\
 & $\alpha=0.01$ & 18.57 & 26.51 & 33.91 & 36.63 & 46.25 & 50.33 & 61.07 & 60.65 & 66.67 & 72.80 \\
 & $\alpha=0.05$ & 26.07 & 39.93 & 41.95 & 49.42 & 51.88 & 54.25 & 70.47 & 61.29 & 68.89 & 71.20 \\
 & $\alpha=0.1$ & 30.00 & 40.94 & 46.55 & 47.09 & 50.63 & 61.44 & \uline{71.14} & 65.16 & 71.85 & 76.80 \\
 & $\alpha=0.5$ & 33.21 & 44.63 & 54.02 & 53.49 & \textbf{63.13} & \uline{62.75} & \uline{71.14} & 69.03 & 63.70 & 72.00 \\
 & $\Delta$ & \textcolor{applegreen}{+1.79} & \textcolor{applegreen}{+4.02} & \textcolor{applegreen}{+1.72} & \textcolor{downred}{-1.17} & \textcolor{downred}{-1.87} & \textcolor{applegreen}{+1.30} & \textcolor{applegreen}{+5.37} & \textcolor{applegreen}{+3.87} & 0.00 & \textcolor{downred}{-2.40} \\
\hline
\multirowcell{30}{\rotatebox{90}{Bloom-7b1}} & Vanilla & \uline{56.43} & \textbf{73.49} & 72.41 & 65.12 & 68.75 & \textbf{77.12} & 78.52 & 80.65 & 77.04 & 87.20 \\
 & TF-IDF & 40.36 & 56.04 & 62.64 & 61.04 & 65.63 & 71.24 & 76.51 & 76.13 & 77.04 & 84.80 \\
 & Keywords & 38.93 & 53.02 & 62.64 & 59.88 & 65.63 & 67.32 & 75.84 & 74.19 & 71.85 & 77.60 \\
 & Summary & 32.50 & 49.66 & 60.34 & 56.40 & 64.38 & 71.90 & 74.50 & 71.61 & 74.07 & 77.60 \\
 & SelCon & 53.21 & 66.11 & 67.24 & \uline{65.70} & 65.00 & 71.90 & 74.50 & 78.71 & 77.04 & 83.20 \\
 & LLMLingua & \textbf{57.86} & \uline{72.82} & \uline{72.99} & \textbf{67.44} & 68.75 & 74.50 & 75.84 & 81.94 & 78.52 & 88.00 \\
 & Ours & \cellcolor{markcolor}54.64 & \cellcolor{markcolor}69.80 & \cellcolor{markcolor}\textbf{73.56} & \cellcolor{markcolor}65.12 & \cellcolor{markcolor}\uline{71.25} & \cellcolor{markcolor}\textbf{77.12} & \cellcolor{markcolor}\textbf{82.55} & \cellcolor{markcolor}\textbf{83.23} & \cellcolor{markcolor}\uline{82.22} & \cellcolor{markcolor}\textcolor{bestcolor}{\textbf{91.20}} \\
 & $\alpha=0.01$ & 22.14 & 35.57 & 43.68 & 50.58 & 64.38 & 61.44 & 71.14 & 70.97 & 71.85 & 82.40 \\
 & $\alpha=0.05$ & 31.07 & 48.66 & 59.77 & 59.88 & 68.13 & 68.63 & 73.83 & 74.84 & 80.74 & 86.40 \\
 & $\alpha=0.1$ & 36.43 & 51.68 & 59.20 & 59.88 & 70.63 & 71.90 & 75.17 & 77.42 & \textbf{82.96} & \uline{89.60} \\
 & $\alpha=0.5$ & 42.86 & 61.74 & 66.67 & 62.79 & \textbf{73.75} & 75.82 & \uline{79.19} & \uline{82.58} & 81.48 & 85.60 \\
 & $\Delta$ & \textcolor{downred}{-1.79} & \textcolor{downred}{-3.69} & \textcolor{applegreen}{+1.15} & 0.00 & \textcolor{applegreen}{+2.50} & 0.00 & \textcolor{applegreen}{+4.03} & \textcolor{applegreen}{+2.58} & \textcolor{applegreen}{+5.18} & \textcolor{applegreen}{+4.00} \\
\hline
\multirowcell{30}{\rotatebox{90}{Llama-2-chat-13b}} & Vanilla & 51.78 & 60.40 & 56.32 & 61.04 & 59.38 & 54.24 & 69.80 & 74.84 & 79.26 & 84.80 \\
 & TF-IDF & 48.57 & 59.01 & 63.79 & 62.21 & 65.63 & \uline{73.20} & 77.18 & 78.71 & 77.78 & 82.40 \\
 & Keywords & 36.43 & 52.01 & 55.17 & 58.72 & 58.13 & 62.75 & 68.46 & 72.26 & 69.63 & 74.40 \\
 & Summary & 27.86 & 44.63 & 46.55 & 54.07 & 48.13 & 46.41 & 56.38 & 65.16 & 65.93 & 83.20 \\
 & SelCon & 55.00 & \uline{69.13} & 63.79 & 67.44 & 65.00 & 69.28 & 73.15 & 78.71 & 80.00 & 86.40 \\
 & LLMLingua & \uline{58.93} & 68.79 & 63.22 & \uline{71.51} & 65.63 & 68.63 & 72.48 & \uline{80.00} & \textbf{81.48} & \uline{88.00} \\
 & Ours & \cellcolor{markcolor}\textbf{59.64} & \cellcolor{markcolor}\textbf{69.46} & \cellcolor{markcolor}\textbf{69.54} & \cellcolor{markcolor}\textbf{72.67} & \cellcolor{markcolor}\textbf{73.75} & \cellcolor{markcolor}\uline{73.20} & \cellcolor{markcolor}\textbf{81.21} & \cellcolor{markcolor}\textbf{82.58} & \cellcolor{markcolor}\textbf{81.48} & \cellcolor{markcolor}\textbf{89.60} \\
 & $\alpha=0.01$ & 30.00 & 41.61 & 44.83 & 54.65 & 56.25 & 66.01 & 71.14 & 68.39 & 73.33 & 85.60 \\
 & $\alpha=0.05$ & 36.79 & 53.69 & 55.74 & 56.40 & 62.50 & 64.71 & 73.83 & 72.26 & 76.30 & 83.20 \\
 & $\alpha=0.1$ & 43.93 & 61.41 & 62.64 & 63.37 & 65.63 & 70.59 & 75.17 & 75.48 & 77.03 & 84.80 \\
 & $\alpha=0.5$ & 55.00 & 68.46 & \uline{68.97} & 69.19 & \uline{70.00} & \textbf{77.12} & \uline{80.54} & 78.06 & 77.04 & 82.40 \\
 & $\Delta$ & \textcolor{applegreen}{+7.86} & \textcolor{applegreen}{+9.06} & \textcolor{applegreen}{+13.22} & \textcolor{applegreen}{+11.63} & \textcolor{applegreen}{+14.37} & \textcolor{applegreen}{+18.96} & \textcolor{applegreen}{+11.41} & \textcolor{applegreen}{+7.74} & \textcolor{applegreen}{+2.22} & \textcolor{applegreen}{+4.80} \\
\hline
\multirowcell{30}{\rotatebox{90}{Llama-3.1-8B-Instruct}} & Vanilla & 61.43 & 76.17 & 72.41 & 74.42 & 70.63 & 79.74 & 83.22 & 82.58 & 86.67 & 89.60 \\
 & TF-IDF & 51.07 & 66.78 & 64.37 & 70.93 & 70.63 & 73.20 & 78.52 & 81.94 & 81.48 & 75.20 \\
 & Keywords & 51.79 & 56.04 & 62.07 & 63.37 & 61.88 & 69.28 & 73.83 & 76.13 & 77.04 & 81.60 \\
 & Summary & 38.93 & 56.38 & 55.75 & 65.12 & 50.63 & 67.32 & 68.46 & 67.10 & 65.19 & 72.00 \\
 & SelCon & 68.21 & 76.85 & 76.44 & 76.16 & 75.63 & 79.08 & 84.56 & \uline{87.10} & 82.22 & \textcolor{bestcolor}{\textbf{91.20}} \\
 & LLMLingua & \uline{73.21} & \textcolor{bestcolor}{\textbf{85.23}} & 78.74 & 77.91 & 76.88 & \uline{84.31} & 83.89 & 86.45 & 88.15 & \textcolor{bestcolor}{\textbf{91.20}} \\
 & Ours & \cellcolor{markcolor}\textcolor{bestcolor}{\textbf{73.93}} & \cellcolor{markcolor}80.20 & \cellcolor{markcolor}\textcolor{bestcolor}{\textbf{81.61}} & \cellcolor{markcolor}\textcolor{bestcolor}{\textbf{79.65}} & \cellcolor{markcolor}\textcolor{bestcolor}{\textbf{78.13}} & \cellcolor{markcolor}\uline{84.31} & \textcolor{bestcolor}{\cellcolor{markcolor}\textbf{89.93}} & \cellcolor{markcolor}\textcolor{bestcolor}{\textbf{89.68}} & \cellcolor{markcolor}\textcolor{bestcolor}{\textbf{90.37}} & \cellcolor{markcolor}\textcolor{bestcolor}{\textbf{91.20}} \\
 & $\alpha=0.01$ & 40.71 & 56.71 & 57.47 & 68.02 & 66.88 & 75.82 & 85.23 & 84.52 & 86.67 & 89.60 \\
 & $\alpha=0.05$ & 50.36 & 67.45 & 67.24 & 74.42 & 73.75 & 80.39 & 87.25 & 85.16 & 88.15 & 88.00 \\
 & $\alpha=0.1$ & 55.00 & 70.81 & 72.99 & 77.91 & \textcolor{bestcolor}{\textbf{78.13}} & \textcolor{bestcolor}{\textbf{84.97}} & \uline{89.26} & 85.81 & \uline{89.63} & 88.80 \\
 & $\alpha=0.5$ & 67.50 & \uline{80.54} & \uline{79.31} & \uline{79.07} & \textcolor{bestcolor}{\textbf{78.13}} & 83.66 & \uline{89.26} & 85.16 & 86.67 & 89.60 \\
 & $\Delta$ & \textcolor{applegreen}{+12.50} & \textcolor{applegreen}{+4.03} & \textcolor{applegreen}{+9.20} & \textcolor{applegreen}{+5.23} & \textcolor{applegreen}{+7.50} & \textcolor{applegreen}{+4.57} & \textcolor{applegreen}{+6.71} & \textcolor{applegreen}{+7.10} & \textcolor{applegreen}{+3.70} & \textcolor{applegreen}{+1.60} \\
\hline
\multirowcell{30}{\rotatebox{90}{DeepSeek-V2-Lite}} & Vanilla & 17.86 & 49.33 & 53.45 & 65.70 & 60.63 & \uline{67.32} & 71.15 & \uline{80.65} & 77.04 & 81.60 \\
 & TF-IDF & 7.14 & 7.38 & 11.49 & 14.53 & 23.75 & 24.18 & 18.79 & 27.74 & 31.85 & 32.00 \\
 & Keywords & 6.07 & 16.78 & 21.26 & 24.42 & 35.00 & 42.48 & 40.94 & 41.94 & 42.96 & 53.60 \\
 & Summary & 9.64 & 35.91 & 44.25 & 52.91 & 53.13 & 62.09 & 57.72 & 67.74 & 72.59 & 80.80 \\
 & SelCon & 23.93 & 54.70 & 49.43 & 56.98 & 57.50 & 62.75 & 72.48 & 78.06 & 75.56 & 76.00 \\
 & LLMLingua & 35.00 & \textbf{67.79} & \uline{67.24} & \textbf{72.09} & \uline{66.88} & 66.67 & \textbf{75.17} & \textbf{83.23} & \uline{84.44} & \uline{84.00} \\
 & Ours & \cellcolor{markcolor}\uline{40.36} & \cellcolor{markcolor}\uline{67.45} & \cellcolor{markcolor}\textbf{68.39} & \cellcolor{markcolor}\textbf{72.09} & \cellcolor{markcolor}\textbf{68.75} & \cellcolor{markcolor}\textbf{67.97} & \cellcolor{markcolor}\uline{73.83} & \cellcolor{markcolor}\uline{80.65} & \cellcolor{markcolor}\textbf{85.19} & \cellcolor{markcolor}\textbf{88.80} \\
 & $\alpha=0.01$ & 23.21 & 37.92 & 35.63 & 47.67 & 41.88 & 47.06 & 58.39 & 56.77 & 59.26 & 68.00 \\
 & $\alpha=0.05$ & 27.50 & 44.97 & 41.38 & 50.58 & 41.88 & 47.06 & 55.03 & 57.42 & 64.44 & 56.80 \\
 & $\alpha=0.1$ & 33.21 & 48.32 & 45.98 & 49.42 & 46.88 & 56.21 & 57.72 & 59.35 & 65.93 & 76.80 \\
 & $\alpha=0.5$ & \textbf{43.57} & 55.70 & 47.13 & 54.65 & 53.13 & 63.40 & 69.80 & 76.13 & 79.26 & 78.40 \\
 & $\Delta$ & \textcolor{applegreen}{+22.50} & \textcolor{applegreen}{+18.12} & \textcolor{applegreen}{+14.94} & \textcolor{applegreen}{+6.39} & \textcolor{applegreen}{+8.12} & \textcolor{applegreen}{+0.65} & \textcolor{applegreen}{+2.68} & 0.00 & \textcolor{applegreen}{+8.15} & \textcolor{applegreen}{+7.20} \\
\hline
\multirowcell{30}{\rotatebox{90}{Qwen3-32B}} & Vanilla & 27.50 & 29.53 & 32.76 & 34.88 & 24.38 & 30.72 & 21.48 & 28.39 & 23.70 & 24.80 \\
 & TF-IDF & 14.29 & 14.77 & 17.82 & 21.51 & 26.25 & 18.30 & 23.49 & 29.03 & 36.30 & 32.00 \\
 & Keywords & 22.86 & 26.85 & 22.99 & 27.33 & 27.50 & 28.10 & 34.23 & 38.06 & 32.59 & 45.60 \\
 & Summary & 22.86 & 28.52 & 30.46 & 29.65 & 29.38 & 33.99 & 28.19 & 35.48 & 34.07 & 48.00 \\
 & SelCon & 21.07 & 26.51 & 26.44 & 26.74 & 31.25 & 29.41 & 42.95 & 34.19 & 41.48 & 47.20 \\
 & LLMLingua & \textbf{30.71} & \textbf{30.54} & \textbf{36.78} & \uline{36.05} & 28.13 & 29.41 & 32.21 & 44.52 & 45.19 & 54.40 \\
 & Ours & \cellcolor{markcolor}\uline{28.92} & \cellcolor{markcolor}\textbf{30.54} & \cellcolor{markcolor}\uline{36.20} & \cellcolor{markcolor}\textbf{36.62} & \cellcolor{markcolor}30.00 & \cellcolor{markcolor}\uline{35.29} & \cellcolor{markcolor}\textbf{43.62} & \cellcolor{markcolor}\uline{45.81} & \cellcolor{markcolor}\textbf{54.81} & \cellcolor{markcolor}\textbf{58.40} \\
 & $\alpha=0.01$ & 11.79 & 19.80 & 25.86 & 24.42 & \textbf{33.13} & 29.41 & 37.58 & 41.29 & 37.78 & 40.80 \\
 & $\alpha=0.05$ & 15.36 & 25.50 & 29.31 & 27.33 & \uline{31.88} & 31.37 & 40.27 & \textbf{52.26} & \textbf{54.81} & \uline{55.20} \\
 & $\alpha=0.1$ & 18.21 & 29.87 & 31.61 & 30.81 & 28.75 & \textbf{35.95} & \textbf{43.62} & 42.58 & 52.59 & 51.20 \\
 & $\alpha=0.5$ & 25.36 & 27.18 & 33.91 & 34.88 & 29.38 & 30.72 & 36.91 & 34.84 & 40.00 & 40.80 \\
 & $\Delta$ & \textcolor{applegreen}{+1.42} & \textcolor{applegreen}{+1.01} & \textcolor{applegreen}{+3.44} & \textcolor{applegreen}{+1.74} & \textcolor{applegreen}{+5.62} & \textcolor{applegreen}{+4.57} & \textcolor{applegreen}{+22.14} & \textcolor{applegreen}{+17.42} & \textcolor{applegreen}{+31.11} & \textcolor{applegreen}{+33.60} \\
\hline

\end{tabular}
\end{adjustbox}
\end{table*}

\begin{table*}[h]
\tiny
\centering
\caption{Accuracy (\texttt{Acc} $\uparrow$) comparison the EntityQuestions dataset. The symbols' definitions are same as Table~\ref{tab:aca_PopQA}.}
\label{tab:aca_EQQA}
\renewcommand{\arraystretch}{0.3}
\begin{adjustbox}{width=0.85\linewidth}
\begin{tabular}{c|c|*{10}{c}}
\hline
LLMs & \diagbox{${C}^{'}$}{${K}$} & 1 & 2 & 3 & 4 & 5 & 6 & 7 & 8 & 9 & 10 \\
\hline
\multirowcell{25}{\rotatebox{90}{GPT-Neo-1.3B}} & Vanilla & 47.24 & 60.31 & 58.45 & 56.95 & \uline{60.25} & \uline{62.31} & 60.34 & \uline{65.09} & \textbf{66.92} & \textbf{71.68} \\
 & TF-IDF & 21.27 & 28.32 & 32.71 & 29.15 & 35.15 & 40.20 & 33.52 & 37.28 & 36.15 & 38.94 \\
 & Keywords & 22.50 & 35.66 & 37.27 & 36.61 & 43.10 & 46.73 & 46.93 & 44.38 & 33.85 & 45.13 \\
 & Summary & 22.29 & 34.09 & 36.46 & 35.93 & 41.84 & 32.16 & 36.87 & 40.83 & 43.85 & 47.49 \\
 & SelCon & 21.06 & 25.70 & 29.76 & 29.15 & 31.80 & 29.65 & 30.17 & 35.51 & 40.77 & 30.09 \\
 & LLMLingua & \textbf{51.53} & \textbf{64.16} & \textbf{60.86} & 56.95 & 56.90 & \textbf{65.83} & \uline{62.01} & 58.58 & 57.69 & \uline{66.37} \\
 & Ours & \cellcolor{markcolor}\uline{50.92} & \cellcolor{markcolor}\uline{61.36} & \cellcolor{markcolor}\uline{59.79} & \cellcolor{markcolor}\uline{57.29} & \cellcolor{markcolor}\textbf{64.85} & \cellcolor{markcolor}57.79 & \cellcolor{markcolor}60.35 & \cellcolor{markcolor}63.31 & \cellcolor{markcolor}\uline{63.08} & \cellcolor{markcolor}\uline{66.37} \\
 & $\alpha=0.01$ & 19.02 & 30.59 & 40.48 & 42.37 & 40.17 & 44.72 & 55.31 & 53.25 & 53.08 & 58.41 \\
 & $\alpha=0.05$ & 25.97 & 39.69 & 44.77 & 49.49 & 52.72 & 53.27 & 59.78 & 63.91 & 55.38 & 59.29 \\
 & $\alpha=0.1$ & 28.63 & 46.33 & 50.94 & 53.56 & 58.58 & 59.30 & \textbf{63.69} & \textbf{66.27} & 58.46 & 60.18 \\
 & $\alpha=0.5$ & 37.01 & 51.05 & 54.69 & \textbf{57.63} & 55.23 & 56.78 & 55.87 & 56.21 & 56.15 & 59.29 \\
 & $\Delta$ & \textcolor{applegreen}{+3.68} & \textcolor{applegreen}{+1.05} & \textcolor{applegreen}{+1.34} & \textcolor{applegreen}{+0.34} & \textcolor{applegreen}{+4.60} & \textcolor{downred}{-4.52} & \textcolor{applegreen}{+0.01} & \textcolor{downred}{-1.78} & \textcolor{downred}{-3.84} & \textcolor{downred}{-5.31} \\
\hline
\multirowcell{25}{\rotatebox{90}{GPT-Neo-2.7B}} & Vanilla & \textbf{54.40} & \uline{64.86} & 65.42 & 64.75 & 67.78 & 69.35 & 71.51 & 68.64 & \uline{72.31} & \uline{73.45} \\
 & TF-IDF & 30.88 & 33.39 & 49.33 & 43.73 & 51.04 & 55.28 & 60.34 & 57.99 & 60.00 & 66.37 \\
 & Keywords & 29.65 & 41.78 & 46.92 & 48.14 & 49.79 & 56.28 & 55.87 & 59.17 & 58.46 & 54.87 \\
 & Summary & 21.06 & 35.14 & 39.14 & 35.93 & 42.26 & 47.74 & 39.66 & 44.38 & 50.00 & 44.25 \\
 & SelCon & 24.74 & 28.67 & 35.92 & 32.20 & 37.66 & 38.69 & 44.13 & 41.42 & 43.08 & 30.09 \\
 & LLMLingua & 54.19 & \textbf{65.56} & 63.81 & 62.71 & \textbf{70.71} & 66.33 & 70.95 & 68.64 & 67.69 & 69.91 \\
 & Ours & \cellcolor{markcolor}\uline{54.21} & \cellcolor{markcolor}62.94 & \cellcolor{markcolor}\textbf{69.71} & \cellcolor{markcolor}\uline{66.78} & \cellcolor{markcolor}\textbf{70.71} & \cellcolor{markcolor}\textbf{71.36} & \cellcolor{markcolor}\textbf{74.86} & \cellcolor{markcolor}\uline{71.60} & \cellcolor{markcolor}\textbf{73.85} & \cellcolor{markcolor}\textbf{76.99} \\
 & $\alpha=0.01$ & 20.45 & 33.22 & 48.53 & 48.47 & 50.21 & 50.75 & 69.27 & 62.13 & 60.77 & 69.91 \\
 & $\alpha=0.05$ & 30.67 & 44.76 & 57.64 & 60.00 & 58.16 & 62.81 & 68.72 & \uline{71.60} & 63.85 & 72.57 \\
 & $\alpha=0.1$ & 36.20 & 52.27 & 59.59 & 62.71 & 63.60 & 63.82 & \uline{72.07} & 71.01 & 67.69 & 68.14 \\
 & $\alpha=0.5$ & 48.46 & 60.14 & \uline{68.10} & \textbf{68.14} & 69.04 & \uline{70.35} & 71.51 & \textbf{73.96} & 71.54 & \uline{73.45} \\
 & $\Delta$ & \textcolor{downred}{-0.19} & \textcolor{downred}{-1.92} & \textcolor{applegreen}{+4.29} & \textcolor{applegreen}{+2.03} & \textcolor{applegreen}{+2.93} & \textcolor{applegreen}{+2.01} & \textcolor{applegreen}{+3.35} & \textcolor{applegreen}{+2.96} & \textcolor{applegreen}{+1.54} & \textcolor{applegreen}{+3.54} \\
\hline
\multirowcell{25}{\rotatebox{90}{OPT-1.3b}} & Vanilla & \textbf{56.24} & \textbf{66.78} & 65.68 & \textbf{65.42} & \uline{73.22} & \uline{67.84} & 70.39 & 69.82 & \uline{75.38} & \uline{70.80} \\
 & TF-IDF & 32.92 & 41.26 & 47.99 & 45.42 & 56.90 & 45.73 & 46.37 & 45.56 & 46.82 & 53.98 \\
 & Keywords & 31.29 & 37.59 & 52.82 & 53.22 & 60.67 & 59.30 & 60.34 & 59.76 & 63.84 & 64.60 \\
 & Summary & 27.40 & 40.03 & 46.65 & 50.17 & 51.46 & 53.77 & 54.75 & 50.89 & 61.54 & 55.75 \\
 & SelCon & 27.61 & 31.64 & 37.53 & 37.29 & 42.26 & 38.69 & 43.58 & 49.11 & 46.15 & 38.05 \\
 & LLMLingua & \uline{54.81} & \uline{66.26} & \uline{66.49} & 57.29 & 68.20 & 64.82 & \uline{73.18} & \uline{71.01} & 63.08 & 69.91 \\
 & Ours & \cellcolor{markcolor}53.41 & \cellcolor{markcolor}62.76 & \cellcolor{markcolor}\textbf{69.71} & \cellcolor{markcolor}63.73 & \cellcolor{markcolor}\textbf{76.15} & \cellcolor{markcolor}\textbf{70.85} & \cellcolor{markcolor}\textbf{74.86} & \cellcolor{markcolor}\textbf{75.15} & \cellcolor{markcolor}\textbf{76.15} & \cellcolor{markcolor}\textbf{73.45} \\
 & $\alpha=0.01$ & 21.47 & 35.66 & 49.06 & 51.86 & 50.63 & 56.78 & 63.69 & 62.72 & 63.85 & 61.95 \\
 & $\alpha=0.05$ & 30.06 & 42.66 & 54.42 & 60.00 & 61.09 & 60.80 & 67.04 & 65.09 & 65.38 & 64.60 \\
 & $\alpha=0.1$ & 34.97 & 51.57 & 61.13 & 59.66 & 64.44 & 64.32 & 68.16 & 67.46 & 68.46 & 64.60 \\
 & $\alpha=0.5$ & 46.63 & 58.74 & 65.15 & \uline{65.08} & 64.85 & 66.33 & 67.04 & 68.64 & 68.46 & 68.14 \\
 & $\Delta$ & \textcolor{downred}{-2.83} & \textcolor{downred}{-4.02} & \textcolor{applegreen}{+4.03} & \textcolor{downred}{-1.69} & \textcolor{applegreen}{+2.93} & \textcolor{applegreen}{+3.01} & \textcolor{applegreen}{+4.47} & \textcolor{applegreen}{+5.33} & \textcolor{applegreen}{+0.77} & \textcolor{applegreen}{+2.65} \\
\hline
\multirowcell{25}{\rotatebox{90}{OPT-2.7b}} & Vanilla & \uline{57.46} & \uline{70.80} & 72.12 & 71.53 & \textbf{76.99} & \textbf{78.39} & \textbf{77.65} & \textbf{79.29} & \textbf{82.31} & 79.64 \\
 & TF-IDF & 34.97 & 41.96 & 50.67 & 51.19 & 63.60 & 64.82 & 63.13 & 64.50 & 68.46 & 62.83 \\
 & Keywords & 33.95 & 43.01 & 52.82 & 59.32 & 65.69 & 62.81 & 69.83 & 67.46 & 70.77 & 73.45 \\
 & Summary & 27.20 & 42.13 & 48.79 & 52.20 & 53.56 & 49.25 & 49.16 & 52.07 & 52.31 & 48.67 \\
 & SelCon & 30.06 & 35.14 & 43.70 & 42.37 & 45.19 & 44.22 & 46.37 & 48.52 & 49.23 & 43.36 \\
 & LLMLingua & \textbf{57.87} & \uline{70.80} & \textbf{75.34} & \uline{74.24} & 74.90 & 73.37 & 69.83 & 68.64 & 67.69 & \uline{80.53} \\
 & Ours & \cellcolor{markcolor}56.62 & \cellcolor{markcolor}\textbf{71.50} & \cellcolor{markcolor}\uline{74.80} & \cellcolor{markcolor}\textbf{76.61} & \cellcolor{markcolor}\uline{76.57} & \cellcolor{markcolor}\textbf{78.39} & \cellcolor{markcolor}71.51 & \cellcolor{markcolor}71.01 & \cellcolor{markcolor}72.31 & \cellcolor{markcolor}\textbf{82.30} \\
 & $\alpha=0.01$ & 25.15 & 36.19 & 47.99 & 56.27 & 55.23 & 60.80 & 68.16 & 68.64 & 72.31 & 69.91 \\
 & $\alpha=0.05$ & 34.76 & 45.45 & 57.10 & 66.10 & 64.02 & 70.85 & 69.27 & 71.60 & \uline{74.62} & 72.57 \\
 & $\alpha=0.1$ & 39.47 & 54.72 & 58.71 & 66.10 & 69.87 & 69.35 & \uline{73.18} & 71.01 & \uline{74.62} & 77.88 \\
 & $\alpha=0.5$ & 53.37 & 65.21 & 69.44 & \uline{74.24} & 74.48 & 71.36 & 69.27 & \uline{72.78} & 72.31 & 74.34 \\
 & $\Delta$ & \textcolor{downred}{-0.84} & \textcolor{applegreen}{+0.70} & \textcolor{applegreen}{+2.68} & \textcolor{applegreen}{+5.08} & \textcolor{downred}{-0.42} & 0.00 & \textcolor{downred}{-6.14} & \textcolor{downred}{-8.28} & \textcolor{downred}{-10.00} & \textcolor{applegreen}{+2.66} \\
\hline
\multirowcell{25}{\rotatebox{90}{Bloom-560m}} & Vanilla & \textbf{48.26} & \textbf{56.47} & \uline{53.62} & \textbf{53.22} & 57.32 & 60.80 & \textbf{54.75} & \textbf{59.17} & \textbf{61.53} & \textbf{61.95} \\
 & TF-IDF & 26.18 & 27.27 & 31.37 & 28.14 & 41.00 & 35.18 & 38.55 & 43.79 & 36.15 & 39.82 \\
 & Keywords & 24.74 & 35.31 & 41.29 & 41.69 & 48.54 & 46.23 & \uline{54.19} & 47.33 & 42.31 & 46.90 \\
 & Summary & 21.68 & 34.97 & 43.70 & 40.00 & 45.19 & 50.75 & 51.40 & 46.75 & 46.92 & 51.33 \\
 & SelCon & 16.77 & 19.23 & 25.20 & 20.68 & 25.10 & 30.15 & 34.08 & 36.69 & 34.62 & 34.51 \\
 & LLMLingua & \uline{44.38} & \uline{54.90} & \textbf{56.03} & \uline{48.47} & \textbf{59.41} & \textbf{64.82} & 50.84 & \uline{57.40} & 52.31 & 60.18 \\
 & Ours & \cellcolor{markcolor}42.97 & \cellcolor{markcolor}52.27 & \cellcolor{markcolor}53.35 & \cellcolor{markcolor}47.12 & \cellcolor{markcolor}\uline{59.00} & \cellcolor{markcolor}\uline{64.32} & \cellcolor{markcolor}50.28 & \cellcolor{markcolor}56.21 & \cellcolor{markcolor}\uline{60.77} & \cellcolor{markcolor}59.29 \\
 & $\alpha=0.01$ & 17.59 & 23.25 & 30.56 & 31.19 & 35.56 & 37.69 & 42.46 & 46.15 & 37.69 & 55.75 \\
 & $\alpha=0.05$ & 23.31 & 30.42 & 34.32 & 36.27 & 43.10 & 39.70 & 52.51 & 55.62 & 47.69 & 56.64 \\
 & $\alpha=0.1$ & 29.24 & 36.19 & 39.14 & 39.66 & 47.70 & 43.22 & 53.63 & 55.03 & 48.46 & \uline{61.06} \\
 & $\alpha=0.5$ & 36.40 & 41.08 & 44.77 & 47.80 & 49.37 & 48.74 & 52.21 & 53.25 & 50.00 & 60.18 \\
 & $\Delta$ & \textcolor{downred}{-5.29} & \textcolor{downred}{-4.20} & \textcolor{downred}{-0.27} & \textcolor{downred}{-6.10} & \textcolor{applegreen}{+1.68} & \textcolor{applegreen}{+3.52} & \textcolor{downred}{-4.47} & \textcolor{downred}{-2.96} & \textcolor{downred}{-0.76} & \textcolor{downred}{-2.66} \\
\hline
\multirowcell{25}{\rotatebox{90}{Bloom-7b1}} & Vanilla & \textbf{58.28} & \textbf{74.65} & \textbf{74.26} & \textbf{76.61} & 79.91 & \textbf{82.41} & \textbf{75.98} & \textbf{84.62} & \uline{83.08} & \textbf{89.38} \\
 & TF-IDF & 37.63 & 47.03 & 53.08 & 61.36 & 67.36 & 63.32 & 67.04 & 68.05 & 72.31 & 68.14 \\
 & Keywords & 34.56 & 50.17 & 56.57 & 62.71 & 69.04 & 67.34 & 68.72 & 73.96 & 69.23 & 74.33 \\
 & Summary & 28.63 & 40.73 & 49.33 & 51.86 & 55.23 & 64.82 & 57.54 & 65.09 & 66.15 & 66.37 \\
 & SelCon & 27.81 & 36.36 & 42.36 & 43.05 & 46.44 & 46.73 & 43.58 & 53.85 & 51.54 & 46.90 \\
 & LLMLingua & \uline{52.15} & \uline{71.85} & \uline{71.58} & 70.84 & \uline{82.01} & 80.40 & 72.63 & 81.66 & 75.38 & 76.99 \\
 & Ours & \cellcolor{markcolor}51.12 & \cellcolor{markcolor}71.50 & \cellcolor{markcolor}71.31 & \cellcolor{markcolor}\uline{72.88} & \cellcolor{markcolor}\textbf{83.26} & \cellcolor{markcolor}\uline{81.91} & \cellcolor{markcolor}73.74 & \cellcolor{markcolor}\uline{82.25} & \cellcolor{markcolor}\textbf{83.84} & \cellcolor{markcolor}\uline{84.96} \\
 & $\alpha=0.01$ & 23.72 & 34.44 & 46.38 & 49.49 & 48.12 & 55.78 & 62.57 & 66.27 & 67.69 & 71.68 \\
 & $\alpha=0.05$ & 31.29 & 44.76 & 54.96 & 58.64 & 60.25 & 63.82 & 70.95 & 76.92 & 71.54 & 74.34 \\
 & $\alpha=0.1$ & 36.81 & 53.67 & 60.05 & 65.42 & 67.78 & 72.36 & 72.07 & 78.70 & 75.38 & 75.22 \\
 & $\alpha=0.5$ & 50.31 & 60.31 & 62.47 & 70.51 & 74.48 & 77.39 & \uline{74.86} & 79.88 & 72.31 & 72.57 \\
 & $\Delta$ & \textcolor{downred}{-7.16} & \textcolor{downred}{-3.15} & \textcolor{downred}{-2.95} & \textcolor{downred}{-3.73} & \textcolor{applegreen}{+3.35} & \textcolor{downred}{-0.50} & \textcolor{downred}{-2.24} & \textcolor{downred}{-2.37} & \textcolor{applegreen}{+0.76} & \textcolor{downred}{-4.42} \\
\hline
\multirowcell{25}{\rotatebox{90}{Llama-2-chat-13b}} & Vanilla & 54.40 & 71.69 & 71.05 & 73.22 & 79.08 & 76.38 & 74.30 & 72.78 & 71.54 & 79.64 \\
 & TF-IDF & 49.28 & 55.24 & 69.17 & 70.51 & \uline{84.10} & 78.39 & 80.45 & 81.66 & 80.77 & 82.30 \\
 & Keywords & 39.47 & 53.85 & 60.86 & 64.41 & 67.36 & 74.37 & 73.18 & 80.47 & 79.23 & 81.42 \\
 & Summary & 31.29 & 43.71 & 45.30 & 44.07 & 50.63 & 51.76 & 41.90 & 52.66 & 56.15 & 66.37 \\
 & SelCon & 37.63 & 43.18 & 53.35 & 56.27 & 59.41 & 65.33 & 64.25 & 68.05 & 69.23 & 68.14 \\
 & LLMLingua & 59.30 & 73.25 & 72.39 & 80.00 & 81.59 & 78.39 & 79.33 & 80.47 & \uline{81.54} & \uline{84.07} \\
 & Ours & \cellcolor{markcolor}\uline{64.83} & \cellcolor{markcolor}\textbf{76.40} & \cellcolor{markcolor}\textbf{79.36} & \cellcolor{markcolor}\uline{80.34} & \cellcolor{markcolor}\textbf{84.52} & \cellcolor{markcolor}\textbf{84.42} & \cellcolor{markcolor}\textbf{85.47} & \cellcolor{markcolor}\textbf{86.39} & \cellcolor{markcolor}\textbf{86.15} & \cellcolor{markcolor}\textbf{86.72} \\
 & $\alpha=0.01$ & 36.40 & 41.61 & 57.37 & 61.36 & 68.20 & 74.37 & 73.18 & 80.47 & 75.38 & 72.57 \\
 & $\alpha=0.05$ & 45.40 & 55.77 & 68.90 & 71.53 & 80.33 & 77.39 & \uline{81.56} & 83.43 & 79.23 & 80.53 \\
 & $\alpha=0.1$ & 49.69 & 65.03 & 74.26 & 76.61 & 81.59 & 82.41 & \uline{81.56} & \textbf{86.39} & 79.23 & 81.42 \\
 & $\alpha=0.5$ & \textbf{65.64} & \textbf{76.40} & \uline{77.75} & \textbf{82.03} & \uline{84.10} & \uline{82.91} & 81.01 & 81.07 & \uline{81.54} & 81.42 \\
 & $\Delta$ & \textcolor{applegreen}{+10.43} & \textcolor{applegreen}{+4.71} & \textcolor{applegreen}{+8.31} & \textcolor{applegreen}{+7.12} & \textcolor{applegreen}{+5.44} & \textcolor{applegreen}{+8.04} & \textcolor{applegreen}{+11.17} & \textcolor{applegreen}{+13.61} & \textcolor{applegreen}{+14.61} & \textcolor{applegreen}{+7.08} \\
\hline
\multirowcell{25}{\rotatebox{90}{Llama-3.1-8B-Instruct}} & Vanilla & 66.67 & 81.29 & 84.18 & 82.03 & 82.85 & 83.42 & 86.03 & 85.21 & 85.38 & 80.53 \\
 & TF-IDF & 56.65 & 61.19 & 72.39 & 75.93 & 78.66 & 69.85 & 72.07 & 69.82 & 70.00 & 58.41 \\
 & Keywords & 55.62 & 63.81 & 71.31 & 71.53 & 76.15 & 80.40 & 86.03 & 81.66 & 76.15 & 78.76 \\
 & Summary & 39.26 & 48.08 & 54.42 & 55.59 & 60.25 & 61.81 & 55.87 & 60.36 & 60.77 & 69.03 \\
 & SelCon & 41.41 & 50.35 & 61.39 & 59.66 & 59.41 & 68.34 & 63.12 & 62.72 & 68.46 & 61.95 \\
 & LLMLingua & 74.44 & \textcolor{bestcolor}{\textbf{85.31}} & 86.60 & \textcolor{bestcolor}{\textbf{93.56}} & 87.45 & 89.95 & 88.83 & 90.53 & 89.23 & 88.50 \\
 & Ours & \cellcolor{markcolor}\textcolor{bestcolor}{\textbf{76.89}} & \cellcolor{markcolor}\uline{84.44} & \cellcolor{markcolor}\textcolor{bestcolor}{\textbf{91.15}} & \cellcolor{markcolor}89.49 & \cellcolor{markcolor}\textcolor{bestcolor}{\textbf{92.05}} & \cellcolor{markcolor}\textcolor{bestcolor}{\textbf{93.97}} & \cellcolor{markcolor}\textcolor{bestcolor}{\textbf{93.30}} & \cellcolor{markcolor}\textcolor{bestcolor}{\textbf{92.90}} & \cellcolor{markcolor}\textcolor{bestcolor}{\textbf{90.77}} & \cellcolor{markcolor}\textcolor{bestcolor}{\textbf{94.69}} \\
 & $\alpha=0.01$ & 45.40 & 59.62 & 69.44 & 74.58 & 80.75 & 80.90 & 87.71 & 84.02 & 90.00 & 86.73 \\
 & $\alpha=0.05$ & 55.21 & 69.93 & 78.55 & 80.34 & 87.45 & 87.44 & 90.50 & 88.17 & 88.46 & 90.27 \\
 & $\alpha=0.1$ & 59.71 & 75.70 & 83.11 & 84.07 & 87.45 & 86.93 & \uline{92.74} & \uline{91.72} & 88.46 & \uline{92.92} \\
 & $\alpha=0.5$ & \uline{76.48} & 83.91 & \uline{88.74} & \uline{90.51} & \uline{90.79} & \uline{91.96} & 89.39 & 89.94 & \textbf{90.77} & 89.38 \\
 & $\Delta$ & \textcolor{applegreen}{+10.22} & \textcolor{applegreen}{+3.15} & \textcolor{applegreen}{+6.97} & \textcolor{applegreen}{+7.46} & \textcolor{applegreen}{+9.20} & \textcolor{applegreen}{+10.55} & \textcolor{applegreen}{+7.27} & \textcolor{applegreen}{+7.69} & \textcolor{applegreen}{+5.39} & \textcolor{applegreen}{+14.16} \\
\hline
\multirowcell{25}{\rotatebox{90}{DeepSeek-V2-Lite}} & Vanilla & 19.63 & 41.78 & 41.02 & 61.36 & 66.95 & 76.88 & \textbf{74.30} & \textbf{75.73} & \textbf{84.62} & \uline{80.53} \\
 & TF-IDF & 6.13 & 8.74 & 12.06 & 16.95 & 22.18 & 20.60 & 20.11 & 24.85 & 35.38 & 33.63 \\
 & Keywords & 4.29 & 13.11 & 20.91 & 28.47 & 29.29 & 29.15 & 38.55 & 48.52 & 49.23 & 49.56 \\
 & Summary & 9.82 & 33.39 & 41.55 & 42.37 & 49.79 & 56.78 & 56.98 & 63.91 & 63.08 & 70.80 \\
 & SelCon & 15.33 & 31.11 & 37.80 & 38.64 & 48.12 & 47.74 & 46.37 & 55.03 & 55.38 & 54.87 \\
 & LLMLingua & 32.72 & \textbf{60.31} & \textbf{67.56} & \textbf{76.95} & \uline{79.08} & \textbf{82.41} & 68.72 & \uline{73.96} & 83.85 & 78.76 \\
 & Ours & \cellcolor{markcolor}\uline{44.38} & \cellcolor{markcolor}\uline{59.62} & \cellcolor{markcolor}\uline{66.75} & \cellcolor{markcolor}\uline{74.58} & \cellcolor{markcolor}\textbf{80.75} & \cellcolor{markcolor}\textbf{82.41} & \cellcolor{markcolor}67.60 & \cellcolor{markcolor}69.82 & \cellcolor{markcolor}\textbf{84.62} & \cellcolor{markcolor}\textbf{87.61} \\
 & $\alpha=0.01$ & 27.81 & 39.69 & 46.92 & 48.47 & 56.90 & 63.82 & 64.25 & 59.17 & 60.77 & 72.57 \\
 & $\alpha=0.05$ & 34.56 & 46.50 & 53.35 & 56.27 & 51.46 & 59.80 & 57.54 & 60.36 & 60.77 & 69.03 \\
 & $\alpha=0.1$ & 37.01 & 51.92 & 56.30 & 56.27 & 57.74 & 61.81 & 65.36 & 63.91 & 66.15 & 73.45 \\
 & $\alpha=0.5$ & \textbf{46.26} & 59.09 & 62.73 & 58.64 & 58.58 & 74.87 & \uline{69.83} & 68.64 & 77.69 & 78.76 \\
 & $\Delta$ & \textcolor{applegreen}{+24.75} & \textcolor{applegreen}{+17.84} & \textcolor{applegreen}{+25.73} & \textcolor{applegreen}{+13.22} & \textcolor{applegreen}{+13.80} & \textcolor{applegreen}{+5.53} & \textcolor{downred}{-6.70} & \textcolor{downred}{-5.91} & 0.00 & \textcolor{applegreen}{+7.08} \\
\hline
\multirowcell{25}{\rotatebox{90}{Qwen3-32B}} & Vanilla & 29.65 & 28.15 & 34.05 & 28.47 & 25.94 & 32.66 & 18.99 & 17.75 & 23.08 & 23.01 \\
 & TF-IDF & 14.52 & 18.36 & 22.79 & 28.14 & 31.38 & 30.15 & 24.58 & 29.59 & 29.23 & 28.32 \\
 & Keywords & 25.15 & 27.79 & 31.90 & 32.88 & 36.40 & 33.17 & 30.17 & 34.32 & 31.54 & 32.74 \\
 & Summary & 25.36 & 28.32 & 24.66 & 28.47 & 28.03 & 20.60 & 20.11 & 18.34 & 16.92 & 23.89 \\
 & SelCon & 17.79 & 18.53 & 26.54 & 28.14 & 28.87 & 23.12 & 24.58 & 27.22 & 30.00 & 21.24 \\
 & LLMLingua & \textbf{42.74} & \textbf{44.06} & \textbf{47.45} & 43.39 & 46.44 & 39.20 & 34.64 & 40.24 & 35.38 & \uline{45.13} \\
 & Ours & \cellcolor{markcolor}\uline{37.22} & \cellcolor{markcolor}\uline{41.08} & \cellcolor{markcolor}43.17 & \cellcolor{markcolor}\textbf{48.81} & \cellcolor{markcolor}\textbf{50.21} & \cellcolor{markcolor}\textbf{45.23} & \cellcolor{markcolor}\textbf{48.60} & \cellcolor{markcolor}\textbf{44.97} & \cellcolor{markcolor}\textbf{42.31} & \cellcolor{markcolor}\textbf{46.02} \\
 & $\alpha=0.01$ & 20.45 & 26.92 & 32.17 & 36.95 & 43.93 & 40.70 & 41.34 & 31.36 & 33.85 & 43.36 \\
 & $\alpha=0.05$ & 23.31 & 33.39 & 39.95 & 38.31 & 42.26 & 41.71 & 42.46 & \uline{42.60} & \uline{37.69} & 36.28 \\
 & $\alpha=0.1$ & 26.58 & 40.21 & 39.95 & 42.03 & 46.86 & 42.21 & \uline{44.13} & 40.24 & \uline{37.69} & 36.28 \\
 & $\alpha=0.5$ & 36.81 & 40.03 & \uline{43.70} & \uline{45.23} & \uline{48.12} & \uline{43.72} & 34.64 & 27.81 & 34.62 & 38.94 \\
 & $\Delta$ & \textcolor{applegreen}{+7.57} & \textcolor{applegreen}{+12.93} & \textcolor{applegreen}{+9.12} & \textcolor{applegreen}{+20.34} & \textcolor{applegreen}{+24.27} & \textcolor{applegreen}{+12.57} & \textcolor{applegreen}{+29.61} & \textcolor{applegreen}{+27.22} & \textcolor{applegreen}{+19.23} & \textcolor{applegreen}{+23.01} \\
\hline

\end{tabular}
\end{adjustbox}
\end{table*}

\section{Ablation Study}
\label{sec:ablation}

We perform an ablation study to analyze the impact of the hyper-parameter $\alpha$, which controls the significance threshold in the concept pruning process, on the overall performance of our method and the results are shown in Table~\ref{tab:ablation}. This parameter determines which concepts are retained from the AMR graphs based on their entropy values to construct the compressed context. Table~\ref{tab:ablation} shows that the lower values of $\alpha$ overly restrict the retained information, pruning out useful concepts and leading to degraded performance. Conversely, higher $\alpha$ values retain too many concepts, which may introduce noise and reduce compression efficiency.

Based on the aforementioned observation, we set $\alpha = 0.3$ in our method, which represents the optimal trade-off, maximizing the discriminatory power of retained concepts while maintaining a compact and informative context for downstream inference. This tuning contributes significantly to the robustness and effectiveness of our context compression approach in context engineering.

\begin{table*}[htbp]
\centering
\caption{The ablation study results of \texttt{AUC} $\uparrow$. The LLMs' order and symbol definitions are the same as Table~\ref{tab:PopQA}.}
\begin{adjustbox}{width=0.85\linewidth}
\begin{tabular}{c|cccccccccccc}
\hline
Datasets   &    $\alpha$                  & $K$          & G-1.3 & G-2.7 & O-1.3 & O-2.7 & b-560 & b-7b1 & L-13 & L3.1-8 & DS-V2 & Q3-32 \\ \hline \hline
\multirow{18}{*}{\rotatebox{90}{\textbf{PopQA}}} &   \multirow{2}{*}{\rotatebox{0}{\footnotesize{Ours}}} & {$I_s$} & \textbf{600.62} & \textbf{611.43} & \textbf{625.14} & \textbf{648.91} & \textbf{587.98} & \textbf{677.77} & \textbf{678.51} & \textbf{756.44} & \textbf{648.90} & \textbf{356.55}    \\ 
                        
                         &  & {$I_l$} & \textbf{283.54} & \textbf{296.09} & 298.73 & \textbf{308.92} & \textbf{292.74} & \textbf{332.16} & \textbf{326.67} & \textbf{357.74} & \textbf{318.06} & \textbf{191.09}    \\ \cline{2-13}
&   \multirow{4}{*}{\rotatebox{0}{\footnotesize{$0.01$}}} & {$I_s$} & 474.99 & 476.62 & 513.82 & 521.05 & 427.70 & 521.88 & 534.01 & 646.48 & 430.18 & 275.57    \\ 
                        
                         &  & {$I_l$} & 261.01 & 255.40 & 286.18 & 279.83 & 249.96 & 285.88 & 288.66 & 339.13 & 231.95 & 151.76    \\ 
                         &  & {$\Delta_{I_s}$} & \textcolor{downred}{-125.63} &\textcolor{downred}{-134.81} &\textcolor{downred}{-111.32} &\textcolor{downred}{-127.86} &\textcolor{downred}{-160.28} &\textcolor{downred}{-155.89} &\textcolor{downred}{-144.50} &\textcolor{downred}{-109.96} &\textcolor{downred}{-218.72} &\textcolor{downred}{-80.98}    \\ 
                         &  & {$\Delta_{I_l}$} & \textcolor{downred}{-22.53} & \textcolor{downred}{-40.69} & \textcolor{downred}{-12.55} & \textcolor{downred}{-29.09} & \textcolor{downred}{-42.78} & \textcolor{downred}{-46.28} & \textcolor{downred}{-38.01} & \textcolor{downred}{-18.61} & \textcolor{downred}{-86.11} & \textcolor{downred}{-39.33}    \\ \cline{2-13}
&   \multirow{4}{*}{\rotatebox{0}{\footnotesize{$0.05$}}}     & {$I_s$} & 518.36 & 527.80 & 555.57 & 585.21 & 486.72 & 593.21 & 575.42 & 692.99 & 444.91 & 328.01    \\ 
                         
                         &  & {$I_l$} & 268.39 & 268.61 & 290.00 & 302.73 & 263.38 & 306.92 & 296.35 & 344.75 & 228.82 & \underline{190.62}    \\ 
                         &  & {$\Delta_{I_s}$} &\textcolor{downred}{-82.26} &\textcolor{downred}{-83.63} &\textcolor{downred}{-69.57} &\textcolor{downred}{-63.70} &\textcolor{downred}{-101.26} &\textcolor{downred}{-84.56} &\textcolor{downred}{-103.09} &\textcolor{downred}{-63.45} &\textcolor{downred}{-203.99} &\textcolor{downred}{-28.54}    \\ 
                         &  & {$\Delta_{I_l}$} &\textcolor{downred}{-15.15} &\textcolor{downred}{-27.48} &\textcolor{downred}{-8.73} &\textcolor{downred}{-6.19} &\textcolor{downred}{-29.36} &\textcolor{downred}{-25.24} &\textcolor{downred}{-30.32} &\textcolor{downred}{-12.99} &\textcolor{downred}{-89.24} &\textcolor{downred}{-0.47}    \\ \cline{2-13}
&   \multirow{4}{*}{\rotatebox{0}{\footnotesize{$0.1$}}}     & {$I_s$} &  518.36 & 555.59 & 590.69 & 604.98 & 508.20 & 611.86 & 615.68 & 721.41 & 484.82 & \underline{330.48}    \\ 
                         
                         &  & {$I_l$} & 268.39 & \underline{288.02} & \textbf{302.36} & \underline{304.22} & \underline{277.27} & 316.30 & 305.38 & \underline{351.58} & 249.50 & 182.36    \\ 
                         &  & {$\Delta_{I_s}$} &\textcolor{downred}{-82.26} &\textcolor{downred}{-55.84} &\textcolor{downred}{-34.45} &\textcolor{downred}{-43.93} &\textcolor{downred}{-79.78} &\textcolor{downred}{-65.91} &\textcolor{downred}{-62.83} &\textcolor{downred}{-35.03} &\textcolor{downred}{-164.08} &\textcolor{downred}{-26.07}    \\ 
                         &  & {$\Delta_{I_l}$} &\textcolor{downred}{-15.15} &\textcolor{downred}{-8.07} & \textcolor{applegreen}{+3.63} &\textcolor{downred}{-4.70} &\textcolor{downred}{-15.47} &\textcolor{downred}{-15.86} &\textcolor{downred}{-21.29} &\textcolor{downred}{-6.16} &\textcolor{downred}{-68.56} &\textcolor{downred}{-8.73}    \\ \cline{2-13}
&   \multirow{4}{*}{\rotatebox{0}{\footnotesize{$0.5$}}}      & {$I_s$} &  \underline{550.01} & \underline{580.26} & \underline{599.36} & \underline{619.43} & \underline{534.50} & \underline{648.25} & \underline{658.08} & \underline{740.35} & \underline{560.19} & 300.90    \\ 
                         
                         &  & {$I_l$} &   \underline{269.94} & 283.66 & \underline{300.58} & 299.82 & 271.24 & \underline{323.96} & \underline{315.40} & 347.72 & \underline{296.09} & 147.51    \\ 
                         &  & {$\Delta_{I_s}$} &  \textcolor{downred}{-50.61} &\textcolor{downred}{-31.17} &\textcolor{downred}{-25.78} &\textcolor{downred}{-29.48} &\textcolor{downred}{-53.48} &\textcolor{downred}{-29.52} &\textcolor{downred}{-20.43} &\textcolor{downred}{-16.09} &\textcolor{downred}{-88.71} &\textcolor{downred}{-55.65}    \\ 
                         &  & {$\Delta_{I_l}$} &   \textcolor{downred}{-13.60} &\textcolor{downred}{-12.43} & \textcolor{applegreen}{+1.85} &\textcolor{downred}{-9.10} &\textcolor{downred}{-21.50} &\textcolor{downred}{-8.20} &\textcolor{downred}{-11.27} &\textcolor{downred}{-10.02} &\textcolor{downred}{-21.97} &\textcolor{downred}{-43.58}    \\ \hline \hline
\multirow{18}{*}{\rotatebox{90}{\textbf{EntityQuestions}}}  &   \multirow{2}{*}{\rotatebox{0}{\footnotesize{Ours}}} & {$I_s$} & \textbf{546.46} & \textbf{627.41} & \textbf{632.79} & \textbf{662.16} & \textbf{494.45} & \textbf{688.73} & \textbf{738.82} & \textbf{813.86} & \textbf{652.14} & \textbf{406.00}    \\ 
                        
                         &  & {$I_l$} & \textbf{248.82} & \textbf{294.48} & \textbf{298.31} & \textbf{295.18} & \textbf{229.06} & \textbf{323.26} & \textbf{343.58} & \textbf{371.30} & \textbf{307.05} & \textbf{181.50}    \\ \cline{2-13}
&   \multirow{4}{*}{\rotatebox{0}{\footnotesize{$0.01$}}} & {$I_s$} & 398.68 & 468.53 & 475.96 & 513.12 & 321.22 & 478.44 & 586.43 & 693.08 & 490.18 & 319.13    \\ 
                        
                         &  & {$I_l$} & 213.20 & 252.50 & 249.62 & 274.46 & 173.02 & 260.26 & 302.50 & 345.54 & 252.38 & 148.58    \\ 
                         &  & {$\Delta_{I_s}$} & \textcolor{downred}{-147.78} &\textcolor{downred}{-158.88} &\textcolor{downred}{-156.83} &\textcolor{downred}{-149.04} &\textcolor{downred}{-173.23} &\textcolor{downred}{-210.29} &\textcolor{downred}{-152.39} &\textcolor{downred}{-120.78} &\textcolor{downred}{-161.96} &\textcolor{downred}{-86.87}    \\ 
                         &  & {$\Delta_{I_l}$} &\textcolor{downred}{-35.62} &\textcolor{downred}{-41.98} &\textcolor{downred}{-48.69} &\textcolor{downred}{-20.72} &\textcolor{downred}{-56.04} &\textcolor{downred}{-63.00} &\textcolor{downred}{-41.08} &\textcolor{downred}{-25.76} &\textcolor{downred}{-54.67} &\textcolor{downred}{-32.92}    \\ \cline{2-13}
&   \multirow{4}{*}{\rotatebox{0}{\footnotesize{$0.05$}}}     & {$I_s$} & 461.64 & 539.16 & 523.81 & 572.68 & 379.61 & 554.66 & 661.10 & 743.58 & 497.84 & 348.16    \\ 
                         
                         &  & {$I_l$} & 235.35 & 271.86 & 260.21 & \underline{287.21} & 203.99 & 288.49 & 323.18 & 355.98 & 243.08 & \underline{161.74}    \\ 
                         &  & {$\Delta_{I_s}$} &\textcolor{downred}{-84.82} &\textcolor{downred}{-88.25} &\textcolor{downred}{-108.98} &\textcolor{downred}{-89.48} &\textcolor{downred}{-114.84} &\textcolor{downred}{-134.07} &\textcolor{downred}{-77.72} &\textcolor{downred}{-70.28} &\textcolor{downred}{-154.30} &\textcolor{downred}{-57.84}    \\ 
                         &  & {$\Delta_{I_l}$} &\textcolor{downred}{-13.47} &\textcolor{downred}{-22.62} &\textcolor{downred}{-38.10} &\textcolor{downred}{-7.98} &\textcolor{downred}{-25.07} &\textcolor{downred}{-34.77} &\textcolor{downred}{-20.40} &\textcolor{downred}{-15.32} &\textcolor{downred}{-63.97} &\textcolor{downred}{-19.76}    \\ \cline{2-13}
&   \multirow{4}{*}{\rotatebox{0}{\footnotesize{$0.1$}}}     & {$I_s$} &  \underline{501.54} & 564.93 & 554.98 & 596.23 & 408.18 & 601.44 & 692.64 & 766.50 & 534.69 & \underline{364.75}    \\ 
                         
                         &  & {$I_l$} & \underline{248.16} & 276.75 & 268.54 & 292.42 & 209.26 & 299.94 & \underline{329.10} & \underline{362.84} & 263.05 & 161.30    \\ 
                         &  & {$\Delta_{I_s}$} &\textcolor{downred}{-44.92} &\textcolor{downred}{-62.48} &\textcolor{downred}{-77.81} &\textcolor{downred}{-65.93} &\textcolor{downred}{-86.27} &\textcolor{downred}{-87.29} &\textcolor{downred}{-46.18} &\textcolor{downred}{-47.36} &\textcolor{downred}{-117.45} &\textcolor{downred}{-41.25}    \\ 
                         &  & {$\Delta_{I_l}$} &\textcolor{downred}{-0.66} &\textcolor{downred}{-17.73} &\textcolor{downred}{-29.77} &\textcolor{downred}{-2.76} &\textcolor{downred}{-19.80} &\textcolor{downred}{-23.32} &\textcolor{downred}{-14.48} &\textcolor{downred}{-8.46} &\textcolor{downred}{-44.00} &\textcolor{downred}{-20.20}    \\ \cline{2-13}
&   \multirow{4}{*}{\rotatebox{0}{\footnotesize{$0.5$}}}      & {$I_s$} &  491.76 & \underline{613.74} & \underline{581.67} & \underline{632.94} & \underline{435.51} & \underline{633.65} & \underline{720.34} & \underline{798.94} & \underline{592.58} & 355.74    \\ 
                         
                         &  & {$I_l$} &   226.26 & \underline{288.91} & \underline{271.38} & \underline{287.21} & \underline{209.92} & \underline{302.03} & 325.79 & 360.77 & \underline{292.97} & 138.40    \\ 
                         &  & {$\Delta_{I_s}$} &  \textcolor{downred}{-54.70} &\textcolor{downred}{-13.67} &\textcolor{downred}{-51.12} &\textcolor{downred}{-29.22} &\textcolor{downred}{-58.94} &\textcolor{downred}{-55.08} &\textcolor{downred}{-18.48} &\textcolor{downred}{-14.92} &\textcolor{downred}{-59.56} &\textcolor{downred}{-50.26}    \\ 
                         &  & {$\Delta_{I_l}$} &   \textcolor{downred}{-22.56} &\textcolor{downred}{-5.57} &\textcolor{downred}{-26.93} &\textcolor{downred}{-7.97} &\textcolor{downred}{-19.14} &\textcolor{downred}{-21.23} &\textcolor{downred}{-17.79} &\textcolor{downred}{-10.53} &\textcolor{downred}{-14.08} &\textcolor{downred}{-43.10}    \\ \hline
\end{tabular}
\end{adjustbox}
\label{tab:ablation}
\end{table*}

\end{document}